\documentclass[twoside]{article}

\usepackage{xcolor}

\usepackage[accepted]{aistats2024}
\usepackage{graphicx}
\usepackage{graphicx}
\usepackage{amsmath, amssymb, amsthm, mathtools}

\usepackage{times}
\usepackage{float}
\DeclareMathOperator*{\argmin}{argmin}
\DeclareMathOperator*{\argmax}{argmax}

\graphicspath{ {./graphics/} }

\usepackage{thmtools} 
\usepackage{thm-restate}

\makeatletter

\newenvironment{proof*}[1][\proofname]{\par
  \pushQED{\qed}%
  \normalfont \partopsep=\z@skip \topsep=\z@skip
  \trivlist
  \item[\hskip\labelsep
        \itshape
    #1\@addpunct{.}]\ignorespaces
}{%
  \popQED\endtrivlist\@endpefalse
}
\makeatother

\newtheorem{theorem}{Theorem}

\newtheorem{lemma}[theorem]{Lemma}
\newtheorem{cor}[theorem]{Corollary}

\newtheorem{prop}[theorem]{Proposition}

\newtheorem{ass}{Assumption}
\newtheorem{ex}{Example}
\newtheorem{definition}{Definition}

\usepackage[round]{natbib}

\usepackage{hyperref}
\usepackage{tikz}
\usetikzlibrary{arrows}
\usepackage{pgfplots}
\begin{document}

\twocolumn[

\aistatstitle{Strategic Usage in a Multi-Learner Setting}

\aistatsauthor{ Eliot Shekhtman \And Sarah Dean }

\aistatsaddress{ Cornell University \And  Cornell University } ]

\begin{abstract}
    Real-world systems often involve some pool of users choosing between a set of services.  With the increase in popularity of online learning algorithms, these services can now self-optimize, leveraging data collected on users to maximize some reward such as service quality.  On the flipside, users may strategically choose which services to use in order to pursue their own reward functions, in the process wielding power over which services can see and use their data.  Extensive prior research has been conducted on the effects of strategic users in single-service settings, with strategic behavior manifesting in the manipulation of observable features to achieve a desired classification; however, this can often be costly or unattainable for users and fails to capture the full behavior of multi-service dynamic systems.  As such, we analyze a setting in which strategic users choose among several available services in order to pursue positive classifications, while services seek to minimize loss functions on their observations.  We focus our analysis on realizable settings, and show that naive retraining can still lead to oscillation even if all users are observed at different times; however, if this retraining uses memory of past observations, convergent behavior can be guaranteed for certain loss function classes.  We provide results obtained from synthetic and real-world data to empirically validate our theoretical findings.
\end{abstract}

\section{INTRODUCTION}

Machine learning (ML) predictions are widely used in today's world, playing an intermediary role between individuals and services in numerous applications. 
In these settings, predictions rarely come from a single entity---instead, multiple service providers collect data on users and train proprietary models.
While this is happening, individuals concurrently choose among these services, making selections according to their own incentives and proportionately
creating a downstream impact on the data available to each service.
In this broader deployment context, 
services must deploy
learning algorithms that can contend with online data collection and shifting distributions.

An example of this can be found in digital credit services offering small short-term loans to individuals who lack access to conventional banking.
This system consists of multiple services, operating largely independently and in an uncoordinated manner.
Services here generally approve or deny loans using an automated system, and make initial lending decisions on the basis of an applicant's information using ML models~\citep{francis2017digital}.
These models are trained on historical lending decisions and are updated as each service collects more data.
Among other possible factors, individual users in this system are incentivized to select a provider who will approve their loan---selecting among services in order to secure a positive classification.

A large body of work on \emph{strategic classification}~\citep{hardt2016strategic}
studies a model of behavior in which individuals modify their data to achieve positive classifications.
This body of work includes the design of decision rules that are robust to anticipated strategic behavior, as well as algorithms for finding such rules through repeated interactions between decision-makers and individuals.
However, this model of data manipulation fails to capture a straightforward way in which individuals can express their preferences: simply choosing amongst alternative providers.

In this work, we formalize the problem of \emph{strategic usage}, where individuals vary their participation in various services according to a strategic objective.
We study a realizable binary classification setting where individuals seek a positive prediction
and services only obtain data from users who select them.
While the usage decision is relatively straightforward, the
differential access to data present in this setting and the resulting multi-learner dynamics are complex.
We show that when services naively update their models with retraining updates, strategic behavior by individuals can cause non-converging oscillations.
Following this realization, we introduce a novel class of 
retraining updates
that make use of memory, which when used can guarantee the convergence of the learning dynamics to an invariant set regardless of initialization.
These invariant sets furthermore exhibit favorable conditions: services experience zero loss across users, correctly classifying all users who choose to use them, and users whose true label is negative will elect to leave the system entirely.

The paper is organized as follows. We begin in Section~\ref{sec:related-work} by reviewing the body of related work on strategic classification and usage choices.
Our {first contribution} is introduced in Section~\ref{sec:problem-setting},
formalizing the strategic usage problem setting and notation and expanding on user and service updates with the novel addition of memory.
In Section~\ref{sec:theoretical-results} we present our {second contribution}, being a characterization of the resulting learning dynamics.
These results are illustrated in Section~\ref{sec:experiments} with numerical simulations on real and synthetic data, and
in Section~\ref{sec:conclusion} we conclude with a discussion of implications and directions for future work.

\section{RELATED WORK}\label{sec:related-work}

\textbf{Strategic Classification}\quad
Our work is inspired by the setting of \emph{strategic classification}, first proposed by \cite{hardt2016strategic},
in which strategic users manipulate their features in order to receive a positive classification,
and a learner or decision maker is tasked with designing a classifier robust to these manipulations.
Many works in this setting study complexities such as the distinction between gaming and improvement~\citep{kleinberg2020classifiers}, connections to causal inference~\citep{miller2020strategic}, and the social burden~\citep{milli2019social}.
As this form of gaming is distinct from the one we study, we do not attempt a comprehensive review of this vast body of work;
instead, we highlight work that considers phenomena relevant to our usage setting.
One such phenomenon is decision-dependent access to data, which in our setting arises due to user choices between services.
\cite{harris2023strategic} studies an online variant of the strategic classification problem in the presence of ``apple tasting'' or one-sided feedback, in which labels are only collected for data points that are positively classified.
\cite{chien2023algorithmic} refer to this phenomenon as algorithmic censoring and explore its implications.
In contrast to these works, in the strategic usage setting, both features and labels are unavailable to services that a user does not select.
Another phenomenon of interest is repeated interaction between learning algorithms and strategic agents.
\cite{dong2018strategic} present algorithms for an online variant of strategic classification in which data arrives sequentially, each point responding to the currently deployed classifier.
\cite{zrnic2021leads} study the dynamics of repeated interactions between a decision-maker and a strategic population, and show convergence to a unique equilibrium depending on their relative update frequencies.
Interestingly, they show that repeated retraining is sufficient to counteract manipulations when their update frequencies are high enough.
Though we study similar repeated retraining dynamics, our analysis differs in that equilibria are not unique.

\textbf{Endogeneous Distribution Shift}\quad
The dynamics of repeated interactions between learning algorithms and endogenously shifting populations have also been studied more generally.
\emph{Performative prediction}, first introduced by
\citet{perdomo2020performative},
generalizes the setting of strategic classification,
modeling a single learner 
seeking to maximize accuracy subject to an 
underlying decision-dependent data distribution. 
They also study the convergence of repeated retraining.
\cite{narang2022multiplayer,
piliouras2022multi, wood2022stochastic}
study a multi-player scenario, where the data distribution depends on the decisions of multiple learners.
We also consider decision-dependent distributions to model the varying usage of strategic individuals;
however, the mechanics of the dependence violate assumptions necessary to apply previous work in the performative setting, such as distribution smoothness. 
In contrast the unique equilibrium of performative prediction dynamics, our setting requires a careful convergence analysis in terms of invariant sets rather than single points.
The framework of \emph{performative power} introduced by \citet{hardt2022performative} studies the ability of services to influence data distributions.
Interestingly, they show that in the presence of a choice between competing providers, 
individuals have no incentive to perform costly manipulations to their features.

\textbf{Usage Choices}\quad
A largely separate body of work has investigated the impacts of ML by studying user participation choices.
\cite{hashimoto2018fairness,zhang2019group} consider sub-populations choosing whether or not to use a single ML model on the basis of accuracy or performance,
showing that retention dynamics can lead to the exacerbation of disparate representation found in the population.
\cite{ginart2021competing,kwon2022competition,dean2022multi} consider users selecting among various services, also on the basis of model accuracy.
For such non-strategic usage, these works characterize the convergence of a simple repeated retraining dynamic.
We show that when usage decisions are made strategically, naive repeated retraining may fail to converge.
Another line of work considers explicit competition for market share between multiple providers, where the market share is determined by users selecting based on performance or accuracy~\citep{gradwohl2022coopetition,aridor2020competing,jagadeesan2022competition,ben2017best,ben2019regression}.
While these works investigate strategic behavior on the part of services, our focus is on strategic behaviors by users.
Our setting departs from all works mentioned in this subsection in that we model users who seek positive classifications, rather than merely high accuracy.

\section{PROBLEM SETTING}\label{sec:problem-setting}
\newcommand{\x}{x}
\newcommand{\veca}{\pmb{a}}
We study the interactions between $n\in\mathbb N_+$ individuals, which we refer to as \emph{users}, and $m\in\mathbb N_+$ machine learning-based services.
Each user $i\in\{1,\dots,n\}$ is represented by features $\x_i\in\mathcal X$,
which encode the information about user $i$ that is available at decision time.
For example, in a digital credit example, this information may include data about an applicant's mobile phone usage, financial transactions, location information, and social media use, among other details~\citep{francis2017digital}.
Also associated with each user $i$ is a label $y_i\in\{+1,-1\}$ indicating an outcome of interest. We refer to $+1$ as a positive label, which is generally seen as desirable, and $-1$ as a negative label, which is generally seen as undesirable.
Unlike the features, the label is not visible at decision time.
In the simplified digital credit example, the label corresponds to whether an individual has the financial resources to repay a loan on time; however, users need not correspond directly to human individuals. 
For example, in simulation experiments, we use the Banknote Authentication dataset~\citep{misc_banknote_authentication_267}.
In this setting, a user corresponds to a banknote, the features are defined by a processed image of the banknote, and the label is whether or not the banknote is authentic.

Users interact with services.
Each service $j\in\{1,\dots,m\}$ selects a classifier $h_j:\mathcal X\to \{+1,-1\}$ from some model class $\mathcal H$.
This classifier predicts the unseen label of a user, given their features.
As with labels, a positive prediction of $+1$ is seen as desirable, while a negative prediction of $-1$ is seen as undesirable. 
In the digital credit example, a prediction determines whether a credit service offers a loan to an individual.
In the banknote example, a prediction determines whether the financial transaction is accepted.

We study the \emph{realizable} setting, meaning that we assume there exists a classifier $h\in\mathcal H$ that perfectly classifies all users simultaneously, $y_i=h(x_i)$ for $i=1,\dots,m$.
This assumption, formally stated in Assumption~\ref{ass:realizability}, is consistent with modern machine learning practice:  expressive high-parameter model classes and high-dimensional data allow for state-of-the-art dataset interpolation~\citep{zhang2017understanding}.
Several works in strategic classification also use this setting 
\citep{nair2022strategic, ghalme2021strategic}.

\begin{ex}\label{ex:linearmodels}
For a given feature transformation $\varphi:\mathcal X\to\mathbb R^d$, the \emph{linear model class} is defined as\\
$\mathcal H = \Big\{h(x)=\begin{cases}+1 & \theta^\top \varphi(x)> 0\\ -1 & \theta^\top \varphi(x)\leq 0\end{cases} ~~\text{s.t.}~~ \theta\in\mathbb R^d\Big\}.$
\end{ex}

Unlike in a classical supervised machine learning setting, or even in the usual strategic classification setting~\citep{hardt2016strategic,zrnic2021leads}, we do not assume that services have immediate access to data about the users.
Instead, the data observed by services depends on the strategic choices of the users.
The following sections describe the user behavior and the learning updates of the services.

\subsection{Strategic Users}\label{sec:settinguser}

Strategic users seek positive classifications, as these correspond to desirable outcomes.
This desire is independent of the user's true label.
Unlike prior work on strategic classification,
in our setting, users \emph{cannot manipulate their data}, but \emph{can select between different services} and \emph{vary their level of usage}.
Thus, the features $x_i$ and label $y_i$ of each user $i$ are fixed. 
Instead, users strategically adjust their \emph{usage}, denoted by $A_{ij}\in\mathbb R_{+}$ for user $i$ and service $j$.
Each user $i$ selects $m$ usage values $A_{i1},\dots,A_{im}$.
These values are non-negative and real-valued, meaning that the user can modulate the total amount of usage.
For example, in the digital credit setting, usage could correspond to the number or amount of loans.

One component of the strategic user objective is the utility from positive classification.
This utility $u:\mathcal X\times\mathcal H \to \mathbb R$ depends on the user features and the classifier. 
We make the following assumptions about the utility: 
\begin{ass}[User utility]
    \label{ass:util}
    For any $h_1,h_2\in\mathcal H$ and $x \in \mathcal{X}$ such that $h_1(x)=-1$ and $h_2(x)=1$, $u(h_1,x)\leq0<u(h_2,x)$.
\end{ass}
This assumption states the intuition that users seek positive classification.
Furthermore, we allow the utility to depend on the classifier $h$, not merely the individual classification $h(x)$.
This allows users to be sensitive to other qualities of the classifier, such as the margin, i.e. the distance from a negative classification. 
We illustrate this in the following example:

\begin{ex}
\label{ex:utility}
For a linear model class where classifiers are represented by their weights $\theta$, the \emph{0-1 utility} is given by $u(x, \theta) = \mathbf 1\{\theta^\top \varphi(x)> 0\}$.
This models users who care only about whether their classification is positive.
The linear utility is given by 
$u(x, \theta) = \theta^\top \varphi(x)$. This models users who care about their margin. 
\end{ex}

The benefit that a user $i$ receives from a service $j$ depends on both utility and usage:
$A_{ij} u(x_i, h_j)$.

The second component of the user objective depends only on their total usage: $\sum_{j=1}^m A_{ij}$.
In particular, higher total usage corresponds to a lower overall objective value. 
This models an opportunity cost or a penalty for large usages on a per-user basis. 
In the digital credit example, individuals are disincentivized from applying for loans too often.

Putting these components together, the overall strategic objective for a user $i$ is to maximize
\begin{align}
    \sum_{j=1}^m A_{ij} u(x_i, h_j) - \frac{1}{q} \Big(\sum_{j=1}^m A_{ij}\Big)^q\label{eq:user_objective}
\end{align}
where the power $q>1$ sets the opportunity cost to increase superlinearly,
so the marginal utility of usage is non-increasing.
To understand this objective, consider the best response of a user $i$ for fixed services $h_1,\dots,h_m$.
If no service offers a positive classification, then by Assumption~\ref{ass:util}, no utility will be positive, and thus the best response will be zero usage.
If there is a service $j$ offering uniquely maximum utility $u(x_i, h_j)>0$, 
then the best response is to ignore any other service,
i.e., $A_{ik}=0$ for all $k\neq j$.
The best response usage will be $A_{ij}=u(h_j,x_i)^{\frac{1}{1-q}}$, thus depending on the maximal utility in addition to the power $q$.

If several services offer equal and maximal utility to the user, then the best response is not unique.
It includes any configuration of usage distributed over these equivalent services, with fixed total usage.
In other words, users consider only the services offering them a sufficiently positive classification, and their usage magnitude is determined by the magnitude of the utility.

We note that implicit in this behavior is that users have full knowledge of service classifiers.  This is a common assumption in strategic classification literature and can be motivated through openly released classifiers or hearsay. In the imperfect knowledge setting, we add that services observe users that they classify as negative, making the game easier for the services.

\subsection{Learning Updates for Services}

Services deploy classifiers trained on data that they have observed.
We take the number of services to be much smaller than the number of users $m\ll n$.
To each service, the large and fluid user pool is represented as a \emph{distribution} over features and labels.
The distribution observed by a service depends on the usages 
so that the weight on data point $(x_i, y_i)$ for service $j$ is proportional to the usage $A_{ij}$.
Importantly, this means that when the usage is zero, the service has no information about the datapoint.
Users who never choose to use a particular service are essentially invisible to the service.

We model the process of training a classifier by expected loss minimization.
The loss $\ell:\mathcal H\times \mathcal X\times \mathcal Y\to\mathbb R$ quantifies the error of a given classifier on a given data point.
We make the following assumption about the loss function.

\begin{ass}[Loss-utility relationship]
    \label{ass:loss-util}
    For all $h \in \mathcal{H}$, the loss is non-negative, $- y \ell (h, x, y)$ strictly monotonically increases with $u(x, h)$, and there exists $v > 0$ such that $u(x, h) = 0\implies \ell(h, x, y) = v$. 
\end{ass}
For a positive user ($y=+1$), this assumption gives a negative relationship between classification utility and the service's quantification of error: high utility corresponds to low error, so the goals of users and services are aligned.
For a negative user ($y=-1$), it is a positive relationship: high utility corresponds to high error, so the user is at odds with the service.
The requirement of the existence of a $v > 0$ meanwhile implies that zero loss gives a user the same sign utility as their label.

\begin{ex}\label{ex:loss}
For a linear model class where classifiers are represented by their weights $\theta$, the \emph{0-1 loss} is $\ell(\theta, x, y) =\mathbf 1\{\theta^\top \varphi(x)\cdot y > 0\}$, which corresponds to the 0-1 utility.
The \emph{hinge loss} is defined as  $\ell(\theta, x, y) = \max\{1-\theta^\top \varphi(x)\cdot y, 0\}$, which corresponds to the linear utility.
\end{ex}

Recall that as discussed above, realizability means that the model class $\mathcal H$ includes a perfect classifier.
We now formalize this intuition in terms of the loss function.

\begin{ass}[Realizability]
    \label{ass:realizability}
    There exists a classifier $h\in\mathcal H$ such that $\ell(h, x_i, y_i)=0$ for all $i=1,\dots n$.
\end{ass}

\begin{ex}
    For the linear model class and either the 0-1 loss or the hinge loss, the realizability assumption is satisfied as long as the features $\{\varphi(x_1),\dots,\varphi(x_n)\}$ are linearly separable with a strictly positive margin.
\end{ex}
This assumption is equivalent to ensuring that there exists a classifier that achieves zero expected loss on the entire population of users.
For a given user distribution $\mathcal D$, the 
expected loss of a classifier $h$ is $\mathbb E_{x,y\sim\mathcal D}[\ell(h, x, y)]$.
When this distribution is defined for a service $j$ based on usages $A_{1j},\dots,A_{nj}$, it can be simplified\footnote{If $\sum_{k=1}^n A_{kj} = 0$ for some service, we adopt the convention that for all users $i$, 
the fraction ${A_{ij}}/({\sum_{k=1}^n A_{kj}}) = 0$.} to
\begin{align}
    \label{eq:lossmin}
    \textstyle
     \sum_{i=1}^n \frac{A_{ij}}{\sum_{k=1}^n A_{kj}}\ell(h, x_i, y_i)\:.
\end{align}
Therefore, when user distributions are defined by usages, the expected loss depends on the usages as well. 

Many works on strategic classification considering the setting where users manipulate their features consider a principle-agent game between a single service and multiple users~\citep{hardt2016strategic}.
These works focus on showing the existence of a desirable equilibrium, where desirability is defined as correct classifications with respect to true user labels.
In our setting, the realizability assumption implies that such good outcomes are possible.
However, because we do not model omniscient services---they observe data depending on strategic usage---arriving at such an equilibrium through interactions requires a nontrivial analysis of dynamical and transient behavior.

\subsection{Interaction Dynamics}

We study the dynamics of repeated interactions between users and services, 
taking a cue from other lines of work on strategic classification \citep{zrnic2021leads}, endogenous distribution shift \citep{perdomo2020performative}, and usage dynamics \citep{dean2022multi}.
In these dynamics, users act according to their strategic objective, while services update their classifiers by retraining.
We thus index classifiers and usage by time, and denote the state of the dynamics as
\[H^t=(h^t_1,\dots h^t_m),\quad A^t \in\mathbb R^{n\times m}\]

We consider user best response dynamics, so at time $t$,
user $i$ selects a usage to maximize \eqref{eq:user_objective} given models $H^t$.
We allow users to employ any tie-breaking scheme (even a non-deterministic one) when there is not a unique maximizing model.
Consider the
following joint update:
\begin{align}
    \label{eq:userjoint}
    \resizebox{0.9\hsize}{!}{$ \displaystyle   A^{t} \in \argmax_{A\in\mathbb R^{n\times m}_+}\sum_{i=1}^n \left[ \sum_{j=1}^m A_{ij} u(x_i, h^t_j) - \frac{1}{q} \Big[\sum_{j=1}^m A_{ij}\Big]^q \right]$}
\end{align}
We consider this joint update for simplicity of exposition, noting that it is
equivalent to any order of independent user updates, due to the separability of the objective.

We consider services that repeatedly retrain their classifiers.
The most naive retraining approach minimizes the immediate expected loss~\eqref{eq:lossmin} given usages observed $A^{t}$, i.e. over the current user distribution.
Such an approach has been studied in the traditional strategic classification setting~\citep{zrnic2021leads,perdomo2020performative}, where it has been shown to converge to desirable equilibria.
As we will see in Section~\ref{sec:mem},
this \emph{memoryless} retraining is a poor fit for the strategic usage setting.

We therefore consider retraining updates which minimize the weighted sum of prior expected losses; that is, the update to models $H^{t+1}$ considers the expected loss~\eqref{eq:lossmin} induced by $A^t, A^{t-1}, \dots $.
By the linearity of expectation, this is equivalent to minimizing the expected loss over a \emph{distribution with memory}.
Thus, rather than depending on current usage $A^t$, classifiers are selected according to an average loss depending on some memory $M^t\in\mathbb R^{n\times m}_+$.
We define memory as follows, using a discount factor $p\geq 0$:
\begin{align}\label{eq:mem}
    M^{t} &= \frac{A^t}{1 + p} + \frac{p M^{t-1}}{1 + p}
\end{align}
By convention, $M^0=0$.
The memoryless case is $p=0$, where the loss depends only on the current timestep. 
For $p>0$, prior observations are retained, but with discounted weight, so that timesteps further in the past contribute less.
As $p$ increases, the memory is ``stronger'': prior timesteps have more influence on the loss.

Therefore, the retraining dynamics are defined by a service $j$ selecting a classifier to minimize the expected loss defined by the memory $M_t$.
The  simultaneous joint update for services is
\begin{align}
    \label{eq:servicejoint}
    H^{t+1} \in \argmin_{H\in\mathcal H^m} 
    \sum_{j=1}^m \sum_{i=1}^n \frac{M^{t}_{ij}}{\sum_{k=1}^n M^{t}_{kj}}\ell(h_j, x_i, y_i)\:.
\end{align}
Due to the separability of the objective over services, this joint update is equivalent to any order of independent service updates.
We thus consider the simultaneous update for simplicity of exposition.

We allow classifiers to employ many types of tie-breaking schemes when there is not a uniquely optimal classifier;
however, we introduce a requirement of sticky tie-breaking.

\begin{definition}
    \label{def:sticky}
    Let there be an update schema where some $f^{t+1}$ is selected to minimize an expected loss as given by $f^{t+1} \in \argmin_{f \in \mathcal{F}} L^t(f)$.
    The update is called \emph{sticky} when if for a given $f^t$, $L^{t-1}(f^t) = L^t(f^t)$, then it holds that $f^{t+1} = f^t$.
\end{definition}
For example, when there is a norm defined on $\mathcal F$, the minimum-norm update rule is \[f^{t+1} = \argmin_{f \in \mathcal{F}} \|f\|_{\mathcal F}\quad \text{s.t.}\quad f \in \argmin_{g \in \mathcal{F}} L^t(g).\]
This update rule satisfies the stickiness property.  
This requirement is grounded in real-world settings as changing classifiers could often be costly or undesirable, and so when no explicit advantage is derived from replacing a model, it might be more advantageous to retain the old model rather than approve and release a new one.

We consider alternating updates between services and users, where the memory evolves according to~\eqref{eq:mem}.
\begin{ass}
    \label{ass:order}
    Given state $(H^t, A^t)$ at time $t$, the classifier update to $H^{t+1}$ is \emph{sticky} and satisfies~\eqref{eq:servicejoint}, and the usage update to $A^{t+1}$ satisfies~\eqref{eq:userjoint}.
\end{ass}
This means that first the services update $H^t\to H^{t+1}$ based on current usage $A^t$, 
and then the users best-respond $A^{t}\to A^{t+1}$ based on the updated services $H^{t+1}$.
We remark that our analysis techniques could extend to a round-robin of updates
involving varied sequences of service and user updates.
We include a further discussion in the appendix and focus on the joint updates here
for simplicity of exposition and notation.

\section{THEORETICAL RESULTS}\label{sec:theoretical-results}

In this section, we formally introduce our theoretical results concerning the dynamics and convergence of the interaction defined by Assumption~\ref{ass:order}.
We will show that under mild assumptions when the memory is non-zero, 
the system will reach a desirable point
within a finite number of steps.
Here, we define desirability from the perspective of accurate classification.
\begin{definition}
    \label{def:zero-loss-eq}
    A state $(H, A)$ is \emph{zero-loss} if
    all services $j$ satisfy: 1)
    $ A_{ij} \ell(h_j, x_i, y_i)=0$ for all $i$ and 2)
    $u(x_i, h_j) \leq 0$ for all $i$ with $y_i=-1$.
\end{definition}
Zero-loss states are desirable from many perspectives.
The first condition means that all
users with nonzero usage have minimal loss.
This means that all services make accurate classifications on the populations they observe.
For positive users, it means maximal utility and correctly positive classifications.
The second condition means that negative users will receive zero utility and will thus choose not to engage in any service.

Beyond desirability from the perspective of utility and loss,
zero-loss points play an important role in understanding the convergence of the interaction dynamics. 
In Section~\ref{sec:equilibria}, we show that for memory parameter $p>0$, when a state is zero-loss, so are all future states.
Thus the zero-loss property defines a set which is \emph{invariant} under the service retraining and user best response updates.
In Section~\ref{sec:convergence}, we further show that the dynamics will reach a zero-loss state in finite time.

All results are presented under Assumptions~\ref{ass:util},~\ref{ass:loss-util}~\ref{ass:realizability}, and~\ref{ass:order}.

\subsection{Importance of Memory}\label{sec:mem}

While naive repeated retraining can be a successful strategy for strategic classification problems in the context of data manipulation~\citep{zrnic2021leads},
in this section we show that it can catastrophically fail.
\begin{restatable}{prop}{propnomemoscilex}
    \label{prop:nomem-oscil-ex}
    In the memoryless $p=0$ setting, there exist settings in which the state $(H, A)$ never converges.
\end{restatable}

\begin{figure}
\centering
\begin{tikzpicture}
\begin{axis}[
    width=.6\columnwidth,
    height=.6\columnwidth,
    axis lines=middle,
    axis line style={draw=none},
    xmin=-1, xmax=1,
    ymin=-1, ymax=1,
    xtick=\empty, ytick=\empty
]
\addplot [only marks, thick, mark=+] table {
0.5 0.5
};
\addplot [only marks, mark=-] table {
-0.5 0.5
0.5 -0.5
-0.5 -0.5
};
\addplot[samples=2, solid,domain=-1:1,black] coordinates {(0, -1)(0, 1)};
\addplot[samples=2, densely dashed,domain=-1:1,black] coordinates {(-1, 0)(1, 0)};
\addplot[samples=2, densely dotted, domain=-1:1,black] coordinates {(-0.5, 1)(1, -0.5)};
\end{axis}
\end{tikzpicture}
    \caption{Five datapoint example described in the proof of Proposition~\ref{prop:nomem-oscil-ex}. Negative points are represented by $-$ and positive points by $+$, where boldface indicates two overlapping points. The dashed and solid lines represent the oscillating classifiers,  with the dotted line representing a zero-loss classifier.}\label{fig:5point}
\end{figure}
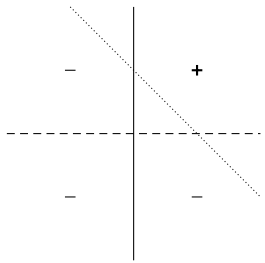

We prove this result by providing an example that leads to oscillations.
We construct a dataset and an initial configuration satisfying all assumptions and show that classifiers and usages will fluctuate indefinitely.
The dataset is illustrated in Figure~\ref{fig:5point}.

\begin{proof*}[Proof Sketch]
    We consider a setting of five users and two classifiers, illustrated in Figure~\ref{fig:5point} and described in detail in Section~\ref{sec:5point}.
    We show that for $p=0$, there are initial conditions that lead to perpendicular oscillating classifiers (dashed and solid in the figure) rather than zero-loss (for example, dotted).
\end{proof*}

In contrast to memoryless updates, when $p>0$, services accumulate knowledge about data distribution over timesteps.
We will show that in our setting, nonzero memory is enough to guarantee convergence to a zero-loss state.
Towards that goal, we first make precise the ability of a service to accumulate knowledge.
\begin{restatable}{lemma}{lemalwayscorrect}
    \label{lem:always-correct}
    For every user service pair $i,j$ such that $M^t_{ij} > 0$, $\ell(h_j^{t+1}, x_i, y_i) = 0$. 
    Therefore, when $p>0$, if $A^{t}_{ij} > 0$ for any timestep $t$, then for all further timesteps $\tau > t$ it must hold that $\ell(h_j^\tau, x_i, y_i) = 0$. 
\end{restatable}

The proof of this result, and the missing proofs of all remaining results, are presented in the appendix.

\subsection{Characterizing Invariance}\label{sec:equilibria}

So far, we have established 
that, in order to avoid oscillations, services must have memory $p>0$ when retraining their classifiers.
Now, we turn to questions of convergence.
However, unlike many related works, we do not study a setting that necessarily admits fixed points, much less unique fixed points.
This occurs because we allow arbitrary user best response updates and assume realizability.
This means that it is plausible for users to vary their usage among several different services indefinitely, so long as those users are positive and the services classify them correctly.
As a result, some care is required
to define the appropriate notion of ``convergence.''
Instead of arguing about fixed points, we turn our attention to an \emph{invariant set} defined by the zero-loss property.
The following proposition formalizes the idea of invariance for zero-loss points.

\begin{restatable}{prop}{propfixediszero}
    \label{prop:fixed-is-zero}
    If a state $(H^t, A^t)$ is zero-loss at time $t$, then all future states are zero-loss for all times $\tau\geq t$. 
\end{restatable}

The proof of this proposition follows from the sticky assumption and users being unable to impart loss on the services without engaging in suboptimal actions. We further establish that the sticky assumption is necessary for classifier tie-breaking in order to guarantee this invariance.

\begin{restatable}{prop}{propnostickynoguarantee}
    \label{prop:no-sticky-no-guarantee}
    Without sticky tie-breaking (Assumption~\ref{ass:order}), regardless of the value of $p$, the existence of a zero-loss equilibrium $(H^t, A^t)$ at a timestep $t$ does not guarantee the existence of zero-loss equilibria for further timesteps $\tau > t$.
\end{restatable}

\begin{proof*}[Proof Sketch]
    We consider a setting where a service is initialized with a model only giving positive usage to positive points, with other negative points never seen.  This may be observed to be a zero-loss state.  At the next timestep, or some future one, the service may feasibly re-sample a new model that gives positive usage to the negative points as they were never observed, resulting in a new state where the zero-loss conditions are violated.
\end{proof*}

\subsection{Convergence}\label{sec:convergence}

Finally, we turn to \emph{convergence}: regardless of the initial state of classifiers and users, we show that the interaction dynamics will lead to a zero-loss point within finite time.

Over the course of interactions between services and users, the services collect data. 
Each time a new user selects a service, the service may respond by updating its classifier---
in particular, a change must occur if the new user is incorrectly classified.
Towards ensuring convergence to a zero-loss point, we hope to show that this process terminates.
The following lemma presents a sufficient condition for reaching a zero-loss point in terms of the usage update.

\begin{restatable}{lemma}{lemmustseenew}
    \label{lem:must-see-new}
    For any timestep $t$ if there exists no values $M^{t-1}_{i,j} = 0$ such that $A^t_{i,j} > 0$, then $(H^t, A^t)$ is zero-loss.
\end{restatable}

The proof of this follows from extending Lemma~\ref{lem:always-correct} to the following timestep.

With this lemma in hand, we are ready to prove the main result,
which shows that services and users will converge to a zero-loss equilibrium in finitely many steps.

\begin{restatable}{theorem}{thmmain}
    \label{thm:main}
    Given nonzero memory $p > 0$, there is a finite time $t \in \mathbb{N}$ after which for all $\tau > t$, $(H^\tau, A^\tau)$ is zero-loss.
\end{restatable}
\begin{proof*}[Proof Sketch]
We first argue that there are only finitely many timesteps in which the condition in Lemma~\ref{lem:must-see-new} can occur.
Therefore, the system must reach a zero-loss point, and by Proposition~\ref{prop:fixed-is-zero}, the classifiers and usages must continue to be zero-loss.
\end{proof*}

\begin{figure*}
    \begin{center}
        \includegraphics[width=.95\textwidth]{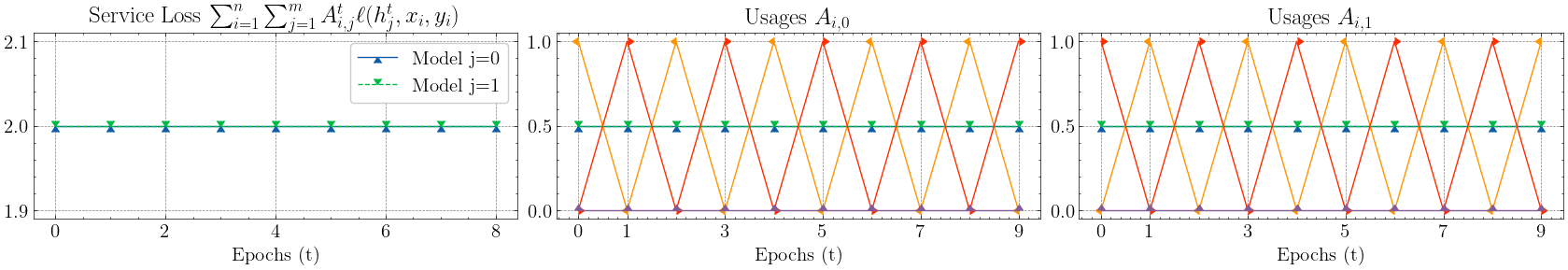} \\
        \includegraphics[width=.95\textwidth]{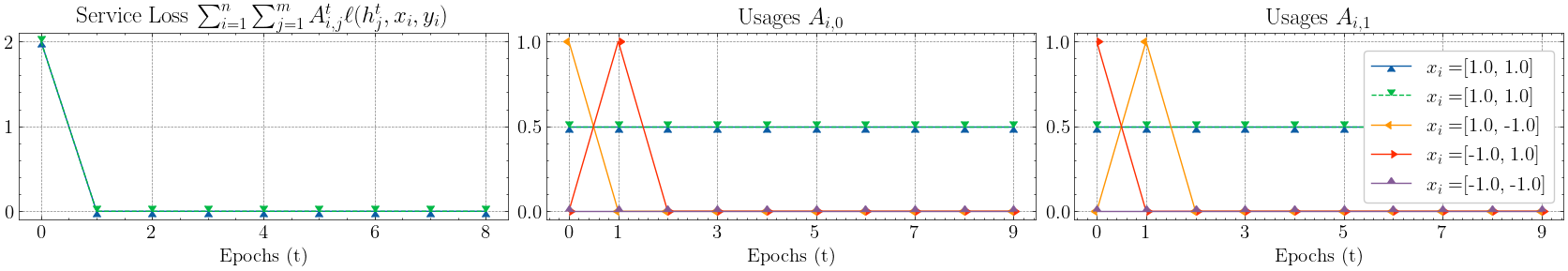}
    \end{center}
    \caption{5-Points dataset; the top three graphs give the $p=0$ case while the bottom three give $p=0.5$.  Service loss is calculated after the user update but before the service update, and usages are displayed for each of the five points with the middle graphs giving the usages for model $j=0$ and the right graphs giving the usages for model $j=1$.}
    \label{fig:5_point}
\end{figure*}
\section{EXPERIMENTS}\label{sec:experiments}

We illustrate our results with experiments of simulated strategic usage behavior.
We first consider a synthetic data setting similar to that found in the proof of Proposition~\ref{prop:nomem-oscil-ex}
to illustrate the importance of memory.
We then perform experiments on real data in the setting of the Banknote Authentication task~\citep{misc_banknote_authentication_267} and the realistic synthetic data of the Bank Account Fraud task~\citep{jesus2022turning}.
These experiments showcase the importance of memory and illustrate the complex and initialization-dependent dynamical behaviors that strategic usage dynamics can induce.

\subsection{Five Points Dataset}
\label{sec:5point}

We begin with a synthetic example with only five points to illustrate oscillation in non-memory cases, with features:
\[\{(1, 1), (1, 1), (-1, 1), (1, -1), (-1, -1)\}\]
and labels $\{1, 1, -1, -1, -1\}$ respectively.
This dataset is clearly separable by linear classifiers with $\varphi(x)=[x, 1]$ and $d=3$ (Example~\ref{ex:linearmodels}). 
We consider users acting strategically according to the linear utility defined in Example~\ref{ex:utility}
with services updating according to the hinge loss defined in Example~\ref{ex:loss}.
We consider $m=2$ services with initial models
$\theta = [1, 0, 0]$ and $\theta = [0, 1, 0]$; the dividing hyperplanes are perpendicular to one another, both giving positive classifications to the coincident positive points, but while model $0$ gives positive classification to $(1, -1)$ the other gives positive classification to $(-1, 1)$.  
It may be observed that this setting is identical to that of the proof of Proposition~\ref{prop:nomem-oscil-ex} and is illustrated in Figure~\ref{fig:5point}.
Note that the fifth user is a negative point who will never choose to use either classifier, and thus will not be seen by them.

For tie-breaking, services choose models by minimizing the norm of $\theta$ (subject to achieving zero loss), while users split usage equally between models that assign equal utility to them.
We run two experiments with this synthetic dataset: one being memoryless with $p=0$ and the other using $p=0.5$.
This demonstrates how the memoryless setting may lead to oscillations while the inclusion of memory ensures convergence.  

Results are presented in Figure~\ref{fig:5_point},
which plots the loss and usage of each service.
In the $p=0$ case, there is clear oscillation in the usages---the users at $(1, -1)$ and $(-1, 1)$ alternate between the two models.  
In contrast, for $p=0.5$, the usages converge after the second epoch.
Perhaps more illuminatingly the loss drops to zero: this indicates that the services have converged to a zero-loss point (Defintion~\ref{def:zero-loss-eq}) and no longer change between updates, unlike in the $p=0$ case where classifiers swap directions each timestep.  
It must also be noted that the usages of the negative users specifically converge to $0$ in the $p>0$ case.
Since the services correctly learn to assign negative classifications to the negative users, even the unseen users at $(-1,-1)$, there is no incentive for nonzero usage from negative users.  
Differing values for $p > 0$ were tested; however, results were similar; relevant figures are present in the appendix. 

\subsection{Banknote Authentication}

We instantiate a semi-synthetic simulation with real-world data coming from the Banknote Authentication dataset \citep{misc_banknote_authentication_267}.  
This dataset involves a binary classification problem to detect whether a banknote is genuine or forged.
Banks are modeled as services and update their forgery-detection classifiers in order to reject forged banknotes while accepting genuine ones.
Meanwhile, users are individuals who seek, for practical purposes, banks that allow them to cash in their notes.
At the same time, banks update their forgery detection models to keep up with trends in the forgery industry.
We remark that our \emph{strategic usage} setting corresponds to the short-term dynamics of banks becoming aware of forged bank notes that are in circulation, 
while the classical feature manipulation setting corresponds to innovations in forgery techniques.

Features are derived from 
images of banknote-like specimens; specifically, they are extracted using a wavelet transform tool, resulting in $\mathcal X = \mathbb R^4$.
Each sample additionally comes with a binary label.
We apply the following preprocessing:
we normalize the mean and variance of the features, and we transform the labels $\{+1, -1\}$.  

We simulate services using scikit-learn support vector classification (SVC) \citep{scikit-learn} with a radial basis function (RBF) kernel using radius $\gamma=1$ and regularization parameter $C=10^{10}$.
This setting corresponds to linear models (Example~\ref{ex:linearmodels}) with an infinite dimensional feature function $\varphi$ trained with hinge loss (Example~\ref{ex:loss}).
The large value of $C$ corresponds to a small regularization weight, which approximates simply selecting the minimum norm model among all loss minimizers.
In the memoryless $p=0$ setting, services have no negative users in their loss objective at various points in time.
This violates the preconditions for the scikit-learn SVC fit function, so in these instances, we preserve weights from the previous timestep.

We initialize this setting by revealing one positive and one negative user at random to each service at a timestep $t=-1$ in order to train models $h^0_j$ for all $j \in \{1, ..., m\}$. 
Random seeds for choosing these users are held constant at $100$ between runs for consistency,
and users tie-break through splitting usage evenly between services that provide equal usage to them.

\begin{figure}[h]
    \begin{center}
        \includegraphics[width=.23\textwidth]{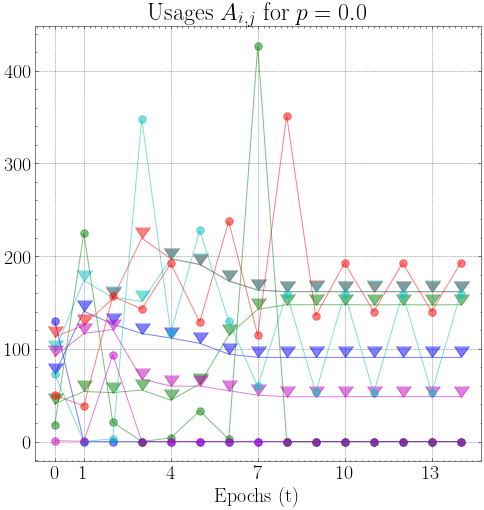} 
        \includegraphics[width=.23\textwidth]{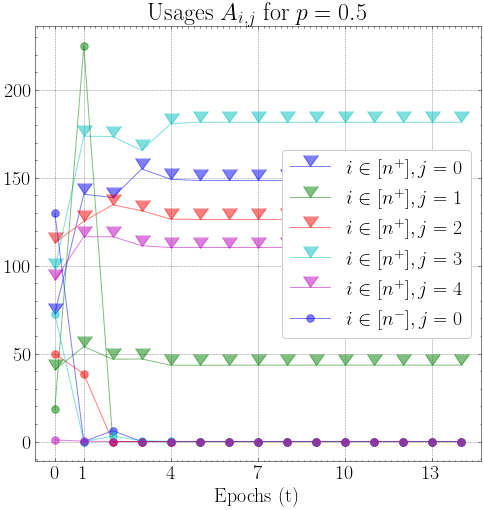}
    \end{center}
    \caption{Banknote Authentication dataset; each graph gives the positive and negative usages of each of the five models; triangle markers above the lines indicate positive usage while below the lines indicate negative, with colors giving which model the line refers to.  The left graph gives the no-memory $p=0$ setting, while the graph on the right gives the $p>0$ setting.  Model order, and hence their colors, are meaningless due to the random initialization. 
    }
    \label{fig:banknote_usages}
\end{figure}

Experiments for various numbers of services, as well as varying the cost power factor $q$, are presented in the appendix.
Figure~\ref{fig:banknote_usages} presents simulation results for $m=5$ services, plotting the total usage of the positive and negative class over time (i.e., of genuine and forged banknotes in circulation).
In the memoryless $p=0$ case, we observe some complex transients until the tenth timestep, after which point we see oscillation between models $j=2$ and $j=3$.
The remaining models stopped observing negative points and henceforth stopped updating;
however, it is important to note that which models converged and which models continued oscillating is initialization dependent: changing the seed resulted in different dynamics, including changing the values that models' usages converged or oscillated around, in addition to changing the numbers of models converging versus oscillating.

In the nonzero memory case $p>0$, we once again observe convergence to a zero-loss point, this time after four timesteps---the longer transient period resulting from the larger number of services that users can choose between. 
At this equilibrium, each model correctly assigns negative classifications to all the negative users, while the positive users are divided between the services. 
The particular configuration of positive users is an arbitrary result of the random initialization; figures depicting the results of other initializations lead to different final configurations and are presented in the appendix.

\subsection{Bank Account Fraud}

We include a third dataset as the Bank Account Fraud (BAF) dataset \citep{jesus2022turning}.  This dataset models the binary classification problem of predicting if a bank account opening is fraudulent or not.  We interpret banks as services providing bank accounts, with users being individuals attempting to open accounts.  As with the banknote authentication setting, this strategic usage setting would provide the precursor to potential feature manipulation which might occur once fraudulent users realize no bank will provide them with utility.  

This dataset suite was designed to stress test the performance and fairness of machine learning models on realistic tabular data generated from a real-world bank account opening fraud dataset.  
None of the datasets fully satisfy our realizability assumption; we selected Variant $V$ because it was the most separable. We filtered the data points to ensure realizability by fitting a soft-margin support vector machine and removing misclassified points.
Each sample has a binary label giving whether the account opening is fraudulent or not, and features include values such as income, customer age, and payment type, exhibiting both real-valued and enumerated types.  We transform non-numerical features into one-hot representations before normalizing the mean and variance of all features, and labels are transformed to $\{+1, -1\}$. 
We present results for a subsample of $10,000$ points due to the large dataset size.
Full details of our implementation and preprocessing can be found in the appendix.

We use Scikit-learn SVC \citep{scikit-learn} to simulate services, with a linear kernel and regularization parameter $C = 10^{10}$.  Services are initialized in a similar manner to the banknote authentication dataset, and a random seed of 100 is held constant between runs for consistency.

\begin{figure}[h]
    \begin{center}
        \includegraphics[width=.23\textwidth]{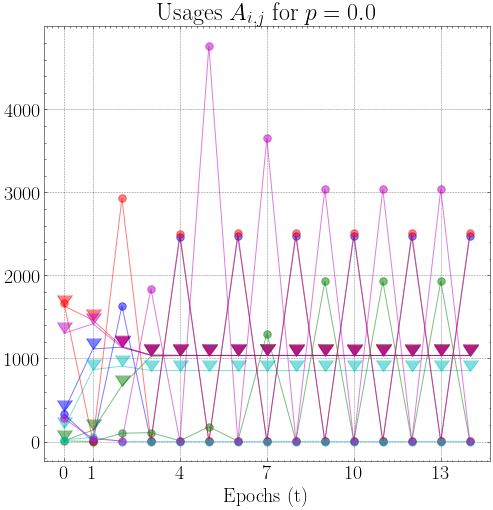} 
        \includegraphics[width=.23\textwidth]{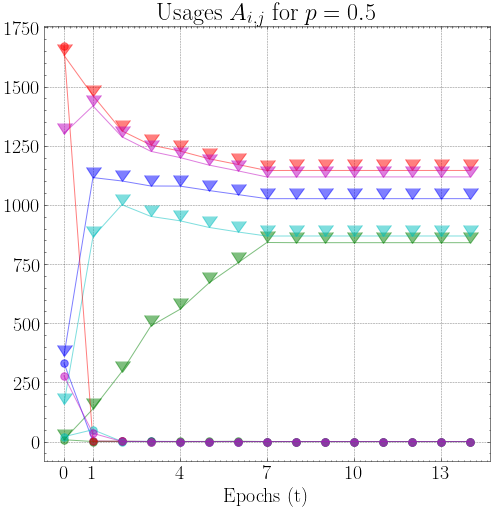}
    \end{center}
    \caption{Bank Account Fraud dataset; each graph gives the positive and negative usages of each of the five models.  Notation is the same as that for Figure~\ref{fig:banknote_usages}.
    }
    \label{fig:baf_usages}
\end{figure}

Experiments for the $p=0$ and $p>0$ can be seen in Figure~\ref{fig:baf_usages}, for $m=5$ services.  We once again observe oscillation and non-zero usage for negative users in the memoryless $p=0$ case, exhibiting the failure of the system to converge to a zero-loss state.  The opposite can be seen in the $p>0$ case, as no change in usage may be observed after the seventh epoch, and all negative users are observed to have left the system.  Further experiments on different initializations may be observed in the appendix.

\section{CONCLUSION \& DISCUSSION}\label{sec:conclusion}

This work focuses on
interaction dynamics between strategic users and multiple learners in an interactive setting.
We formalize \emph{strategic usage}, in which individuals choose to use services strategically in order to pursue a
positive classification. Since usage presents their data to said services, the latter respond by retraining their classifiers, to the potential benefit or detriment of these strategic individuals.
We provide sufficient conditions to guarantee a finite-time equilibrium, including the addition of memory to classifier retraining updates.  
Our work is both an extension of and in contrast to the strategic classification setting, which primarily explores dynamics involving strategic user feature modification in pursuit of goals.
We remark that while several works have raised concern as to the adverse social outcomes entailed by strategic classification \citep{milli2019social, Chen2020StrategicRI, hu2019disparate} our setting ensures that all true positives may receive positive utility at equilibrium.

As the first to study the setting of strategic usage, our work raises several areas for future extension.  
Though the realizable setting is well motivated for modern ML, 
it is natural to analyze the setting in which no single classifier can correctly classify the entire dataset.
Another natural extension is to consider explicit competition between services, rather than retraining blindly, unaware of the existence of models being deployed by other services in the system.  
Similarly, we study un-coordinated and short-term user objectives, 
but an exploration of long-term strategic planning and optimization might give insight into additional real-world phenomena.
Finally, it would be interesting to consider the relationship between a finite user pool and an underlying population-level distribution.

\subsubsection*{Acknowledgements}
This work was supported by NSF CCF 2312774, NSF OAC-2311521, a LinkedIn Research Award, and a gift from Wayfair.

\bibliographystyle{abbrvnat}
\bibliography{references.bib}

\begin{thebibliography}{32}
\providecommand{\natexlab}[1]{#1}
\providecommand{\url}[1]{\texttt{#1}}
\expandafter\ifx\csname urlstyle\endcsname\relax
  \providecommand{\doi}[1]{doi: #1}\else
  \providecommand{\doi}{doi: \begingroup \urlstyle{rm}\Url}\fi

\bibitem[Aridor et~al.(2020)Aridor, Mansour, Slivkins, and Wu]{aridor2020competing}
G.~Aridor, Y.~Mansour, A.~Slivkins, and Z.~S. Wu.
\newblock Competing bandits: The perils of exploration under competition.
\newblock \emph{arXiv preprint arXiv:2007.10144}, 2020.

\bibitem[Ben-Porat and Tennenholtz(2017)]{ben2017best}
O.~Ben-Porat and M.~Tennenholtz.
\newblock Best response regression.
\newblock \emph{Advances in Neural Information Processing Systems}, 30, 2017.

\bibitem[Ben-Porat and Tennenholtz(2019)]{ben2019regression}
O.~Ben-Porat and M.~Tennenholtz.
\newblock Regression equilibrium.
\newblock In \emph{Proceedings of the 2019 ACM Conference on Economics and Computation}, pages 173--191, 2019.

\bibitem[Chen et~al.(2020)Chen, Wang, and Liu]{Chen2020StrategicRI}
Y.~Chen, J.~Wang, and Y.~Liu.
\newblock Strategic recourse in linear classification.
\newblock \emph{ArXiv}, abs/2011.00355, 2020.
\newblock URL \url{https://api.semanticscholar.org/CorpusID:226226668}.

\bibitem[Chien et~al.(2023)Chien, Roberts, and Ustun]{chien2023algorithmic}
J.~Chien, M.~Roberts, and B.~Ustun.
\newblock Algorithmic censoring in dynamic learning systems.
\newblock \emph{arXiv preprint arXiv:2305.09035}, 2023.

\bibitem[Dean et~al.(2022)Dean, Curmei, Ratliff, Morgenstern, and Fazel]{dean2022multi}
S.~Dean, M.~Curmei, L.~J. Ratliff, J.~Morgenstern, and M.~Fazel.
\newblock Multi-learner risk reduction under endogenous participation dynamics.
\newblock \emph{arXiv preprint arXiv:2206.02667}, 2022.

\bibitem[Dong et~al.(2018)Dong, Roth, Schutzman, Waggoner, and Wu]{dong2018strategic}
J.~Dong, A.~Roth, Z.~Schutzman, B.~Waggoner, and Z.~S. Wu.
\newblock Strategic classification from revealed preferences.
\newblock In \emph{Proceedings of the 2018 ACM Conference on Economics and Computation}, pages 55--70, 2018.

\bibitem[Francis et~al.(2017)Francis, Blumenstock, and Robinson]{francis2017digital}
E.~Francis, J.~Blumenstock, and J.~Robinson.
\newblock Digital credit: A snapshot of the current landscape and open research questions.
\newblock \emph{CEGA White Paper}, pages 1739--76, 2017.

\bibitem[Ghalme et~al.(2021)Ghalme, Nair, Eilat, Talgam-Cohen, and Rosenfeld]{ghalme2021strategic}
G.~Ghalme, V.~Nair, I.~Eilat, I.~Talgam-Cohen, and N.~Rosenfeld.
\newblock Strategic classification in the dark.
\newblock In \emph{International Conference on Machine Learning}, pages 3672--3681. PMLR, 2021.

\bibitem[Ginart et~al.(2021)Ginart, Zhang, Kwon, and Zou]{ginart2021competing}
T.~Ginart, E.~Zhang, Y.~Kwon, and J.~Zou.
\newblock Competing ai: How does competition feedback affect machine learning?
\newblock In \emph{International Conference on Artificial Intelligence and Statistics}, pages 1693--1701. PMLR, 2021.

\bibitem[Gradwohl and Tennenholtz(2022)]{gradwohl2022coopetition}
R.~Gradwohl and M.~Tennenholtz.
\newblock Coopetition against an amazon.
\newblock In \emph{International Symposium on Algorithmic Game Theory}, pages 347--365. Springer, 2022.

\bibitem[Hardt et~al.(2016)Hardt, Megiddo, Papadimitriou, and Wootters]{hardt2016strategic}
M.~Hardt, N.~Megiddo, C.~Papadimitriou, and M.~Wootters.
\newblock Strategic classification.
\newblock In \emph{Proceedings of the 2016 ACM conference on innovations in theoretical computer science}, pages 111--122, 2016.

\bibitem[Hardt et~al.(2022)Hardt, Jagadeesan, and Mendler-D{\"u}nner]{hardt2022performative}
M.~Hardt, M.~Jagadeesan, and C.~Mendler-D{\"u}nner.
\newblock Performative power.
\newblock \emph{arXiv preprint arXiv:2203.17232}, 2022.

\bibitem[Harris et~al.(2023)Harris, Podimata, and Wu]{harris2023strategic}
K.~Harris, C.~Podimata, and Z.~S. Wu.
\newblock Strategic apple tasting.
\newblock \emph{arXiv preprint arXiv:2306.06250}, 2023.

\bibitem[Hashimoto et~al.(2018)Hashimoto, Srivastava, Namkoong, and Liang]{hashimoto2018fairness}
T.~Hashimoto, M.~Srivastava, H.~Namkoong, and P.~Liang.
\newblock Fairness without demographics in repeated loss minimization.
\newblock In \emph{International Conference on Machine Learning}, pages 1929--1938. PMLR, 2018.

\bibitem[Hu et~al.(2019)Hu, Immorlica, and Vaughan]{hu2019disparate}
L.~Hu, N.~Immorlica, and J.~W. Vaughan.
\newblock The disparate effects of strategic manipulation.
\newblock In \emph{Proceedings of the Conference on Fairness, Accountability, and Transparency}, FAT* '19, page 259–268, New York, NY, USA, 2019. Association for Computing Machinery.
\newblock ISBN 9781450361255.
\newblock \doi{10.1145/3287560.3287597}.
\newblock URL \url{https://doi.org/10.1145/3287560.3287597}.

\bibitem[Jagadeesan et~al.(2022)Jagadeesan, Jordan, and Haghtalab]{jagadeesan2022competition}
M.~Jagadeesan, M.~I. Jordan, and N.~Haghtalab.
\newblock Competition, alignment, and equilibria in digital marketplaces.
\newblock \emph{arXiv preprint arXiv:2208.14423}, 2022.

\bibitem[Jesus et~al.(2022)Jesus, Pombal, Alves, Cruz, Saleiro, Ribeiro, Gama, and Bizarro]{jesus2022turning}
S.~Jesus, J.~Pombal, D.~Alves, A.~Cruz, P.~Saleiro, R.~P. Ribeiro, J.~Gama, and P.~Bizarro.
\newblock Turning the tables: Biased, imbalanced, dynamic tabular datasets for ml evaluation, 2022.

\bibitem[Kleinberg and Raghavan(2020)]{kleinberg2020classifiers}
J.~Kleinberg and M.~Raghavan.
\newblock How do classifiers induce agents to invest effort strategically?
\newblock \emph{ACM Transactions on Economics and Computation (TEAC)}, 8\penalty0 (4):\penalty0 1--23, 2020.

\bibitem[Kwon et~al.(2022)Kwon, Ginart, and Zou]{kwon2022competition}
Y.~Kwon, A.~Ginart, and J.~Zou.
\newblock Competition over data: how does data purchase affect users?
\newblock \emph{arXiv preprint arXiv:2201.10774}, 2022.

\bibitem[Lohweg(2013)]{misc_banknote_authentication_267}
V.~Lohweg.
\newblock {banknote authentication}.
\newblock UCI Machine Learning Repository, 2013.
\newblock {DOI}: https://doi.org/10.24432/C55P57.

\bibitem[Miller et~al.(2020)Miller, Milli, and Hardt]{miller2020strategic}
J.~Miller, S.~Milli, and M.~Hardt.
\newblock Strategic classification is causal modeling in disguise.
\newblock In \emph{International Conference on Machine Learning}, pages 6917--6926. PMLR, 2020.

\bibitem[Milli et~al.(2019)Milli, Miller, Dragan, and Hardt]{milli2019social}
S.~Milli, J.~Miller, A.~D. Dragan, and M.~Hardt.
\newblock The social cost of strategic classification.
\newblock In \emph{Proceedings of the Conference on Fairness, Accountability, and Transparency}, pages 230--239, 2019.

\bibitem[Nair et~al.(2022)Nair, Ghalme, Talgam-Cohen, and Rosenfeld]{nair2022strategic}
V.~Nair, G.~Ghalme, I.~Talgam-Cohen, and N.~Rosenfeld.
\newblock Strategic representation, 2022.

\bibitem[Narang et~al.(2022)Narang, Faulkner, Drusvyatskiy, Fazel, and Ratliff]{narang2022multiplayer}
A.~Narang, E.~Faulkner, D.~Drusvyatskiy, M.~Fazel, and L.~J. Ratliff.
\newblock Multiplayer performative prediction: Learning in decision-dependent games.
\newblock \emph{arXiv preprint arXiv:2201.03398}, 2022.

\bibitem[Pedregosa et~al.(2011)Pedregosa, Varoquaux, Gramfort, Michel, Thirion, Grisel, Blondel, Prettenhofer, Weiss, Dubourg, Vanderplas, Passos, Cournapeau, Brucher, Perrot, and Duchesnay]{scikit-learn}
F.~Pedregosa, G.~Varoquaux, A.~Gramfort, V.~Michel, B.~Thirion, O.~Grisel, M.~Blondel, P.~Prettenhofer, R.~Weiss, V.~Dubourg, J.~Vanderplas, A.~Passos, D.~Cournapeau, M.~Brucher, M.~Perrot, and E.~Duchesnay.
\newblock Scikit-learn: Machine learning in {P}ython.
\newblock \emph{Journal of Machine Learning Research}, 12:\penalty0 2825--2830, 2011.

\bibitem[Perdomo et~al.(2020)Perdomo, Zrnic, Mendler-D{\"u}nner, and Hardt]{perdomo2020performative}
J.~Perdomo, T.~Zrnic, C.~Mendler-D{\"u}nner, and M.~Hardt.
\newblock Performative prediction.
\newblock In \emph{International Conference on Machine Learning}, pages 7599--7609. PMLR, 2020.

\bibitem[Piliouras and Yu(2022)]{piliouras2022multi}
G.~Piliouras and F.-Y. Yu.
\newblock Multi-agent performative prediction: From global stability and optimality to chaos.
\newblock \emph{arXiv preprint arXiv:2201.10483}, 2022.

\bibitem[Wood and Dall'Anese(2022)]{wood2022stochastic}
K.~Wood and E.~Dall'Anese.
\newblock Stochastic saddle point problems with decision-dependent distributions.
\newblock \emph{arXiv preprint arXiv:2201.02313}, 2022.

\bibitem[Zhang et~al.(2017)Zhang, Bengio, Hardt, Recht, and Vinyals]{zhang2017understanding}
C.~Zhang, S.~Bengio, M.~Hardt, B.~Recht, and O.~Vinyals.
\newblock Understanding deep learning requires rethinking generalization, 2017.

\bibitem[Zhang et~al.(2019)Zhang, Khaliligarekani, Tekin, et~al.]{zhang2019group}
X.~Zhang, M.~Khaliligarekani, C.~Tekin, et~al.
\newblock Group retention when using machine learning in sequential decision making: the interplay between user dynamics and fairness.
\newblock \emph{Advances in Neural Information Processing Systems}, 32, 2019.

\bibitem[Zrnic et~al.(2021)Zrnic, Mazumdar, Sastry, and Jordan]{zrnic2021leads}
T.~Zrnic, E.~Mazumdar, S.~Sastry, and M.~Jordan.
\newblock Who leads and who follows in strategic classification?
\newblock \emph{Advances in Neural Information Processing Systems}, 34:\penalty0 15257--15269, 2021.

\end{thebibliography}

\newpage
\appendix
\onecolumn

\section{Missing Proofs}

In this section, we restate the main theoretical results, include their full proofs, and introduce a handful of auxiliary results.

\subsection{Oscillations and Counter-Examples}

In this section, we present examples of dynamics that do not converge or that are not zero-loss invariant.

\propnomemoscilex*
\begin{proof*}[Proof of Proposition~\ref{prop:nomem-oscil-ex}]
    Let there exist a set of users with features $X = \{(1, 1), (1,1), (-1, 1), (1, -1), (-1, -1)\}$ and labels $Y = \{1, 1, -1, -1, -1\}$, selecting between two services where both services choose models from the linear model class (Example~\ref{ex:linearmodels}) using feature transformation $\phi(x) = (x, 1)$, using linear utility (Example~\ref{ex:utility}) and hinge loss (Examples~\ref{ex:loss}).
    Suppose that the initial models are defined through parameters $\theta_1 = [1, 0, 0]^\top$ and $\theta_2 = [0, 1, 0]^\top$, that retraining updates tie-break by choosing the minimum norm classifier, and that users tie-break by dividing usage equally between services.

    Optimal updates may be calculated by analyzing stable points of repeated gradient updates on the best-response updates found in Equations~\ref{eq:userjoint} and \ref{eq:servicejoint}noting that with $p=0$, $M^t = A^t$ for any timestep $t$.
    This algebra gives us the following update steps:
    \begin{align*}
        \theta^0 &= \begin{bmatrix}
            1 & 0 & 0 \\
            0 & 1 & 0 
        \end{bmatrix}^\top 
        &A^0 &= \begin{bmatrix}
            0.5 & 0.5 & 1 & 0 & 0 \\
            0.5 & 0.5 & 0 & 1 & 0
        \end{bmatrix}^\top \\
        \theta^1 &= \begin{bmatrix}
            0 & 1 & 0 \\
            1 & 0 & 0 
        \end{bmatrix}^\top 
        &A^1 &= \begin{bmatrix}
            0.5 & 0.5 & 0 & 1 & 0 \\
            0.5 & 0.5 & 1 & 0 & 0
        \end{bmatrix}^\top \\
        \theta^2 &= \begin{bmatrix}
            1 & 0 & 0 \\
            0 & 1 & 0 
        \end{bmatrix}^\top 
        &A^2 &= \begin{bmatrix}
            0.5 & 0.5 & 1 & 0 & 0 \\
            0.5 & 0.5 & 0 & 1 & 0
        \end{bmatrix}^\top
    \end{align*}

    It may be observed that $\theta^2 = \theta^0 \not= \theta^1$, and $A^2 = A^0 \not= A^1$. 
    Given the memoryless setting in which $(H^{t+1},A^{t+1})$ depend only on the previous $(H^t,A^t)$, we may conclude that this system will oscillate for all further timesteps and will not reach a fixed state.
\end{proof*}

\propnostickynoguarantee*
\begin{proof*}[Proof of Proposition~\ref{prop:no-sticky-no-guarantee}]
    Let there exist a set of users with features $X = \{(5), (-5), (0)\}$ and labels $Y=\{1, -1, 1\}$, 
    making usage decisions in a system with one service choosing models from the linear model class (Example~\ref{ex:linearmodels}) using feature transformation $\phi(x) = (x, 1)$, using linear utility (Example~\ref{ex:utility}) and hinge loss (Examples~\ref{ex:loss}).
    Suppose that the model is defined through parameters $\theta = [1, -1]^\top$, with retraining updates tie-breaking stochastically.

    Optimal updates may be calculated by analyzing stable points of repeated gradient updates on the best-response updates found in Equations~\ref{eq:userjoint} and \ref{eq:servicejoint}.  
    The first update step can be seen as follows:
    \begin{align*}
        \theta^0 &= \begin{bmatrix}
            1 & -1
        \end{bmatrix}^\top 
        &A^0 &= \begin{bmatrix}
            1 & 0 & 0
        \end{bmatrix}^\top \\
    \end{align*}
    It may be noted that this constitutes a zero-loss equilibrium, as for all users $i$, $A^0_{i} \ell(h^0, x_i, y_i) = 0$, and for the negative point, $\ell(h^0, x_i, y_i) < 1$.  Despite this, it is feasible for the following second update step to occur due to $H$ not being constrained by sticky tie-breaking:
    \begin{align*}
        \theta^1 &= \begin{bmatrix}
            1 & -0.5
        \end{bmatrix}^\top 
        &A^1 &= \begin{bmatrix}
            1 & 0 & 0.5
        \end{bmatrix}^\top \\
    \end{align*}
    Here, it can been that for the point $(0)$, $A^1_{i} \ell(h^1, x_i, y_i) = 0.25$.  As such, state $(H^1, A^1)$ does not constitute a zero-loss equilibrium, proving the claim.
\end{proof*}

\subsection{Invariance}

We next turn to our results about the invariance of zero-loss points.
We begin with lemmas characterizing important properties of the service retraining update.

\lemalwayscorrect*
\begin{proof*}[Proof of Lemma~\ref{lem:always-correct}]
    By the separability of Equation~\ref{eq:servicejoint} on services, we may analyze the best-response update of a particular service $j$.  Let us call $H_j^{t+1}$ the set of models such that the best-response update $h_j^{t+1} \in H_j^{t+1}$.
    \begin{align*}
        H_j^{t+1} \coloneqq \argmin_{h \in \mathcal{H}} \sum_{i=1}^n \frac{M^t_{ij}}{\sum_{k=1}^n M^t_{kj}} \ell(h_j, x_i, y_i)
    \end{align*}
    Since terms of the sum where $M^t_{i,j} = 0$ will simply be zero we may simplify the sum, and furthermore in the interest of brevity let us represent $\frac{M^t_{ij}}{\sum_{k=1}^n M^t_{kj}}$ as $r_{i,j}(M^t)$.
    \begin{align}
        \label{eq:proof-pos-loss-sum}
        &= \sum_{i \in \{i | M_{i,j}^t > 0\}}^n r_{i,j}(M^t) \ell(h_j, x_i, y_i) 
    \end{align}
    By Assumption~\ref{ass:realizability}, we have that there exists an $h^* \in \mathcal{H}$ such that $\forall i \in \{1, ..., n\}$, $\ell(h^*, x_i, y_i) = 0$.  
    By the non-negativity of $\ell$ and $M$ it must hold that $ r_{i,j}(M^t) \ell(h_j, x_i, y_i) \geq 0$ for all $i,j$ pairs, and since substituting $h^*$ into expression~\ref{eq:proof-pos-loss-sum} would return a sum of $0$, it must be that $0$ is the minimum value of the sum.
    For all $h \in \mathcal{H}$ such that $\ell(h, x_i, y_i) > 0$ for some user $i$ where $M_{i,j}^t > 0$, the value of the sum would be greater than $0$ and therefore $h$ wouldn't minimize the sum so $h \not\in H_j^{t+1}$.
    Finally, we may conclude that for all $h_j \in H_j^{t+1}$, $\ell(h, x_i, y_i) \leq 0$ for all $i$ such that $M_{i,j}^t > 0$.

    For the second statement,
    we may first observe that at any timestep $t$, if $A^t_{i,j} > 0$ then by Equation~\ref{eq:mem} and the non-negativity of $M$, $M^t_{i,j} > 0$.
    For the $p>0$ case, if at some timestep $t$ we have that for some user $i$ and service $j$, $M^t_{i,j} > 0$, then it will hold that $M^{\tau-1}_{i,j} > 0$ for all timesteps $\tau > t$.  This may be trivially observed through the non-negativity of $A$ and the memory update given by Equation~\ref{eq:mem}.

    By the conclusions drawn above,
    since $M^{\tau-1}_{i,j} > 0$ we may conclude that $\ell(h_j^\tau, x_i, y_i) \leq 0$.
\end{proof*}

Next, we introduce an auxiliary lemma that formalizes the best response behavior of users.
This lemma makes rigorous the informal discussion at the end of Section~\ref{sec:settinguser}.

\begin{lemma}
    \label{lem:max-dominates}
    For a given user $i$, let there be a set of services $J$ such that 
    $u_0 = u(x_i, h_j^t) > u(x_i, h_k^t)$ for all services $j \in J$ and other services $k \not\in J$.  The user best response at time $t$ must satisfy $u_0^{1/({q-1})} = \sum_{j \in J} A_{i,j}^t > \sum_{k \not\in J} A_{i,k}^t = 0$.
\end{lemma}
\begin{proof*}[Proof of Lemma~\ref{lem:max-dominates}]
    Analyzing the strategic objective for a user $i$ (\ref{eq:user_objective}) and denoting it as $L_i^u$ for convenience, we can begin by seeing that given some set of services $J$ such that $u(x_i, h_{j_1}) = u(x_i, h_{j_2}) = C_0$, if we hold
    $\sum_{j \in J} A_{i,j} = C_1$ constant and
    $A_{i,k}$ constant for all $k \not\in J$ such that $\sum_{k \not\in J} A_{i,k} = C_2$ and $\sum_{k \not\in J} A_{i,k} u(x_i, h_k) = C_3$, then any distribution of usages between services in $J$ doesn't affect the value of the strategic objective.
    \begin{align*}
        L_i^u(A, H) &= \sum_{j=1}^m A_{i,j} u(x_i, h_j) - \frac{1}{q} \left( \sum_{j=1}^m A_{i,j} \right)^q \\
        &= \sum_{j \in J} A_{i,j} u(x_i, h_j) + \sum_{k \not\in J} A_{i,k} u(x_i, h_k) - \frac{1}{q} \left( \sum_{j \in J} A_{i,j} + \sum_{k \not\in J} A_{i,k} \right)^q \\
        &= C_0 \sum_{j \in J} A_{i,j}  + C_3 - \frac{1}{q} \left( \sum_{j \in J} A_{i,j} + C_2 \right)^q \\
        &= C_0 C_1 + C_3 - \frac{1}{q} \left( C_1 + C_2 \right)^q \\
    \end{align*}
    Therefore, without loss of generality, let us say all usage in $J$ is concentrated into a service $j \in J$; since $A_{i,j'} = 0$ for all $j' \in J$, $j' \not= j$ we can drop them from the strategic objective and consider only a service $j$ and services $k \not\in J, k \not= j$.

    Taking the derivative of $L_i^u(A, H)$ with respect to some $A_{i,j}$, we get gradient $\frac{d}{d A_{i,j}} L_i^u (A, H) = u(x_i, h_j) - \left( A_{i,j} + \sum_{k\not=j} A_{i,k} \right)^{q-1}$.  
    Let us take two services $j$ and $k$, such that $u(x_i, h_j) > u(x_i, h_l)$ for all $l \not= j$.  Service $k$ is chosen arbitrarily such that $j \not= k$.
    We shall now enumerate options in a case analysis focusing on the total usage of user $i$.  We shall show that as total usage increases, the model is incentivized to concentrate all usage on the service that provides the highest utility as the derivative of the objective function with respect to other services becomes negative.

    Let us say that $\sum_{l=1}^m A_{i,l} = 0$.  
    For service $j$, we have that $\frac{d}{d A_{i,j}} L_i^u = u(x_i, h_j) > 0$; therefore $A_{i,j}$ is below the optimum.

    Let us say $\sum_{l=1}^m A_{i,l} < u(x_i, h_{k})^{\frac{1}{q-1}}$.
    For service $j$, we have that $\frac{d}{d A_{i,j}} L_i^u = u(x_i, h_j) - \left( \sum_{l=1}^m A_{i,l} \right)^{q-1} > u(x_i, h_j) - u(x_i, h_k) > 0$; therefore $A_{i,j}$ is below the optimum.
    For service $k$, we have that $\frac{d}{d A_{i,k}} L_i^u = u(x_i, h_k) - \left( \sum_{l=1}^m A_{i,l} \right)^{q-1} > u(x_i, h_k) - u(x_i, h_k) = 0$; therefore $A_{i,k}$ is below the optimum.
    
    Let us say $\sum_{l=1}^m A_{i,l} = u(x_i, h_{k})^{\frac{1}{q-1}}$.
    For service $j$, we have that $\frac{d}{d A_{i,j}} L_i^u = u(x_i, h_j) - \left( \sum_{l=1}^m A_{i,l} \right)^{q-1} = u(x_i, h_j) - u(x_i, h_k) > 0$; therefore $A_{i,j}$ is below the optimum.
    For service $k$, we have that $\frac{d}{d A_{i,k}} L_i^u = u(x_i, h_k) - \left( \sum_{l=1}^m A_{i,l} \right)^{q-1} = u(x_i, h_k) - u(x_i, h_k) = 0$; therefore $A_{i,k}$ at an optimum.

    Let us say $u(x_i, h_{k})^{\frac{1}{q-1}} < \sum_{l=1}^m A_{i,l} < u(x_i, h_j)^{\frac{1}{q-1}}$.
    For service $j$, we have that $\frac{d}{d A_{i,j}} L_i^u = u(x_i, h_j) - \left( \sum_{l=1}^m A_{i,l} \right)^{q-1} > u(x_i, h_j) - u(x_i, h_j) = 0$; therefore $A_{i,j}$ is below the optimum.
    For service $k$, we have that $\frac{d}{d A_{i,k}} L_i^u = u(x_i, h_k) - \left( \sum_{l=1}^m A_{i,l} \right)^{q-1} < u(x_i, h_k) - u(x_i, h_k) = 0$; therefore $A_{i,k}$ is above the optimum.

    Let us say $\sum_{l=1}^m A_{i,l} = u(x_i, h_j)^{\frac{1}{q-1}}$.
    For service $j$, we have that $\frac{d}{d A_{i,j}} L_i^u = u(x_i, h_j) - \left( \sum_{l=1}^m A_{i,l} \right)^{q-1} = u(x_i, h_j) - u(x_i, h_j) = 0$; therefore $A_{i,j}$ is at an optimum.
    For service $k$, we have that $\frac{d}{d A_{i,k}} L_i^u = u(x_i, h_k) - \left( \sum_{l=1}^m A_{i,l} \right)^{q-1} = u(x_i, h_k) - u(x_i, h_j) < 0$; therefore $A_{i,k}$ is above the optimum.

    Let us say $\sum_{l=1}^m A_{i,l} > u(x_i, h_j)^{\frac{1}{q-1}}$.
    For service $j$, we have that $\frac{d}{d A_{i,j}} L_i^u = u(x_i, h_j) - \left( \sum_{l=1}^m A_{i,l} \right)^{q-1} < u(x_i, h_j) - u(x_i, h_j) = 0$; therefore $A_{i,j}$ is above the optimum.
    For service $k$, we have that $\frac{d}{d A_{i,k}} L_i^u = u(x_i, h_k) - \left( \sum_{l=1}^m A_{i,l} \right)^{q-1} < u(x_i, h_k) - u(x_i, h_j) < 0$; therefore $A_{i,k}$ is above the optimum.

    From all of this, we can see that the stable equilibrium holds that $A_{i,j} =  u(x_i, h_j)^{\frac{1}{q-1}}$, and that for all services $k \not= j$, $A_{i,k} = 0$ because of the non-negativity of $A$.
\end{proof*}

\begin{cor}
    \label{cor:zero-util}
    For any user $i$ and service $j$ pair such that at timestep $t$, $u(x_i, h_j^t) \leq 0$, then $A_{i,j}^t = 0$.
\end{cor}
\begin{proof*}[Proof of Corollary~\ref{cor:zero-util}]
    By Lemma~\ref{lem:max-dominates}, we have that if there exists some service $k$ such that $u(x_i, h_k^t) > u(x_i, h_j^t)$ then $A_{i,j}^t = 0$.  Let us analyze the case where for all users $k$, $u(x_i, h_j^t) = 0 \geq u(x_i, h_k^t)$.
    Once again taking the derivative of the strategic objective for a user $i$ (\ref{eq:user_objective}) with respect to $A_{i,j}^t$ and denoting it as $\frac{d}{d A_{i,j}^t} L_i^u(A^t, H^t)$ for convenience, we can see that if $\sum_{k=1}^m A_{i,k}^t > 0$ then $\frac{d}{d A_{i,j}^t} L_i^u(A^t, H^t) < 0$.  This indicates that the optimum value of $A_{i,j}^t$ is $0$.
\end{proof*}

Finally, we prove the main invariance result.
\propfixediszero*
\begin{proof*}[Proof of Proposition~\ref{prop:fixed-is-zero}]

    We prove the proposition by showing that
    if a state $(H^t, A^t)$ is a zero-loss equilibrium, then $(H^{t+1}, A^{t+1})$ is also a zero-loss equilibrium.

    We begin by arguing that $H^{t+1}=H^t$.
    Consider the retraining objective for service $j$ at $t$: $ \sum_{i=1}^n \frac{M^t_{i,j}}{\sum_{k=1}^n M^t_{k,j}} \ell (h_j^t, x_i, y_i)$
    By Lemma~\ref{lem:always-correct}, we know that $\ell (h_j^{t}, x_i, y_i) = 0$ for all users $i$ and services $j$ such that $M^{t-1}_{i,j} > 0$.
    By zero-loss condition 1 we have that 
    $\ell (h_j^{t}, x_i, y_i) = 0$ for all users $i$ and services $j$ such that $A^t_{i,j} > 0$.
    Thus by the definition of memory (\ref{eq:mem}), it must be that $\ell (h_j^{t}, x_i, y_i) = 0$ for all users $i$ and services $j$ such that $M^{t}_{i,j} > 0$.
    We conclude that $h_j^t$ achieves zero retraining loss, and therefore by the definition of sticky tie-breaking (\ref{def:sticky}), $h^{t+1}_j=h^t_j$ for all services $j$.
    This implies that the zero-loss condition 2 holds: $0 \geq u(x_i, h_j^t) = u(x_i, h_j^{t+1})$.

    We next argue that if $A^t$ is zero loss, so is $A^{t+1}$.
    First, consider negative users $i$ with $y_i=-1$.
    By the zero-loss condition 2 shown in the previous paragraph $u(x_i, h_j^t) \leq 0$.
    Therefore, by Corollary~\ref{cor:zero-util} the best response is $A_{ij}^{t+1}=0$ for all $j$ and all negative users $i$, thus ensuring that zero-loss condition 1 holds for negative users $i$.

    Now, consider positive users $i$ with $y_i=+1$.
    Because $H^{t+1}=H^t$ the user best-response objective remains the same. 
    By monotonicity of $\ell$, we have that there exists some value $v'$ such that if $\ell(h, x, +1) = 0$, $u(x, h) = v'$.
    As we know that $\ell(h_j^t, x_i, y_i) = 0$ for all $A^t_{i,j} > 0$, we have that for all $A^t_{i,j} > 0$, $u(x_i, h_j^t) = v_{ij}'$.
    Let us say that there exists some positive user $i$ and service $j$ such that $\ell(h_j^{t+1}, x_i, y_i) > 0$, meaning $u(x_i, h_j^{t+1}) < v_{ij}'$.  
    By Lemma~\ref{lem:max-dominates}, this implies that $A^{t+1}_{i,j} = 0$.  
    As such, for all $i,j$, if $\ell(h_j^{t+1}, x_i, y_i) > 0$ then $A^{t+1}_{i,j} = 0$; therefore, $A^{t+1} \ell(h_j^{t+1}, x_i, y_i) = 0$ for all users $i$ and services $j$.  This satisfies zero-loss condition 1 for positive users.
    
    Thus we have shown that if a state $(H^t, A^t)$ is a zero-loss equilibrium, then $(H^{t+1}, A^{t+1})$ is also a zero-loss equilibrium.
    Through induction, this guarantees that the state $(H^\tau,A^\tau)$ is a zero-loss equilibrium for all timesteps $\tau > t$.
\end{proof*}
\subsection{Convergence}
With invariance results in hand, we can now prove the main convergence result.
\lemmustseenew*
\begin{proof*}[Proof of Lemma~\ref{lem:must-see-new}]
    Lets say that there exists some timestep $t$ such that there exists no values $M^{t-1}_{i,j} = 0$ such that $A^t_{i,j} > 0$.  

    By Lemma~\ref{lem:always-correct} and by the non-negativity of $\ell$, this means that $\ell (h_j^t, x_i, y_i) = 0$ for all users $i$ and services $j$ such that $M^{t-1}_{i,j} > 0$.  This gives us that for all users $i$ and services $j$, $\frac{M^{t-1}_{i,j}}{\sum_{k=1}^n M^{t-1}_{k,j}} \ell (h_j^t, x_i, y_i) = 0$ as either $M^{t-1}_{i,j} = 0$ or $\ell (h_j^t, x_i, y_i) = 0$.
    Since $A^t_{i,j} = 0$ when $M^{t-1}_{i,j} = 0$ and when $M^{t-1}_{i,j} > 0$ it holds that $\ell (h_j^t, x_i, y_i) = 0$, we can similarly conclude that for all $i,j$, $A^t_{i,j} \ell (h_j^t, x_i, y_i) = 0$.
    This satisfies the first condition of zero-equilibrium states.

    Let us define an $L^u_i$ as follows:
    \begin{align*}
        L^u_i(A, H) = \frac{1}{q} \left( \sum_{j=1}^m A_{i,j} \right)^q - \sum_{j=1}^m A_{i,j} u(x_i, h_j)
    \end{align*}
    We know that for all $M^{t-1}_{i,j} > 0$, $\ell(h_j^t, x_i, y_i) = 0$ by Lemma~\ref{lem:always-correct} and the non-negativity of $\ell$.  This indicates that for all $i$ such that $y_i = -1$, if $M^{t-1}_{i,j} > 0$ then $u(x_i, h_j^t) \leq 0$.  
    By Corollary~\ref{cor:zero-util}, this gives that $A^t_{i,j} = 0$.  
    Since $A^t_{i,j} = 0$ if $M^{t-1}_{i,j} = 0$ as well, this indicates that for all services $j$, for all users $i$ such that $y_i = -1$, $A^t_{i,j} = 0$.

    Let us say for contradiction that for some user $i$ such that $y_i = -1$, for service $j = \argmax_j u(x_i, h_j^t)$ it holds that $u(x_i, h_j^t) > 0$.
    By Lemma~\ref{lem:max-dominates}, it must hold that $A_{i,j}^t > 0$.  This poses a contradiction, and therefore we may state that for all $i$ such that $y_i = -1$, for all $j$, $u(x_i, h_j^t) \leq 0$.
\end{proof*}
\thmmain*
\begin{proof*}[Proof of Theorem~\ref{thm:main}]
    We may analyze each timestep $t$ as either a timestep in which there exists some user-service pair $i,j$ such that $A^t_{i,j} > 0$ and $M_{i,j}^{t-1} = 0$, or in which no such pair exists.  There are only $n \times m$ such $i,j$ pairs and as such a maximum of $nm$ timesteps where the first condition is satisfied.  In the contrary, at any step $t$ where the condition isn't satisfied, by Lemma~\ref{lem:must-see-new}, we have shown that we have reached a fixed point.
    Therefore, there is a maximum of $nm$ timesteps before the conditions are reached to reach a zero-loss equilibrium.  
    By Proposition~\ref{prop:fixed-is-zero}, this indicates that for all timesteps $\tau \geq nm$, $(H^\tau, A^\tau)$ constitutes a zero-loss equilibrium.
\end{proof*}

We construct examples to indicate the linear dependence of convergence time on $n$ and $m$.
Let us define the model class as $h(x) = \mathrm{sign}(x + \theta)$, with utility as $u(x, h) = \min\{x + \theta, 1\}$ and loss as $\ell(h, x, y) = \max\{0, 1 - y(x + \theta)\}$.  We set $q=2$ and $p> 0$.
User tie-breaking is done by stochastically selecting one service to use, and services tie-break by choosing the minimum change classifier between timesteps.
\begin{ex}
    Let there be $n$ evenly spaced positive users, with features: 
    \[
        \{ (0), (-0.7), (-1.4), ..., (-0.7)(2-n), (-0.7)(1-n) \}:.
    \]
    If $m=1$ service is instantiated with model $\theta^0 = 0.5$, at every timestep $t$, user $t$ will receive positive utility and elect for positive usage, pushing the classifier to $\theta^t = 1-x_t$.  This will result in $nm=n$ total timesteps before convergence.
\end{ex}
\begin{ex}
    Let there be $n=1$ user such that $X = \{(0)\}$ and $Y = \{-1\}$.
    If $m$ services are instantiated with models $\theta^0_j = j+1$, at every timestep $t$, some service $j$ will receive usage from the user, pushing the classifier to $\theta^t_j = -1$, at which point it'll never receive positive usage again.  This will result in $nm=m$ total timesteps before convergence.
\end{ex}

\subsection{Round Robin Updates}

It might not be realistic that users and services update on the same schedule: while prior proofs assume that services conduct one synchronous update based on current usage and users synchronously best respond once based on the updated services at every timestep, it could be that services only update every several years while users may reallocate usage yearly.
Additionally, users and services don't necessarily update synchronously, and one might conduct several updates at a time while others only do one.
In this setting, a timestep stops being a feasible metric of progression; instead, we generalize to the concept of a round. 

Instead of there being one joint user and one joint service update as in a timestep, we generalize rounds to users and services updating asynchronously and differing numbers of times.  We maintain three constraints on this system.  First, rounds are divided into alternating user update periods and service update periods, such that only users or services are updating at a time.  Second, each round must contain at least one user period and one service period.  Finally, each user and each service undergoes at least one best response update in each period.

\begin{prop}
    \label{prop:round-robin}
    Given nonzero memory $p > 0$, there there are a finite number of rounds $r \in \mathbb{N}$ after which for all $\rho > r$, $(H^\rho, A^\rho)$ is zero-loss.
\end{prop}
\begin{proof*}[Proof of Proposition~\ref{prop:round-robin}]
    We shall prove this by showing that a round is functionally equivalent to a set of timesteps that assume stochastic user tie-breaking.  Since we have that a zero-loss point will be reached after a finite number of timesteps by Theorem~\ref{thm:main}, this will imply the existence of a zero-loss point after a finite number of rounds.

    As we have already analyzed the result of a service best responding to a set of users, and a user best responding to a set of services, let us analyze the change when services and users undergo multiple consecutive updates.  

    Given a service $j$ at update $k>0$, let us denote $h^k_j$ as the best response to memory $M^k_j$ updating on usages $A$.
    By the memory update (\ref{eq:mem}), we have that for all users $i$, $M^k_{i,j} = 0$ if and only if $M^{k+1}_{i,j} = 0$.
    By Lemma~\ref{lem:always-correct}, we have that $M_{i,j}^{k+1} \ell(h_j^k, x_i, y_i) = 0$ for all users $i$.  As $h^k_j$ achieves a value of zero for the objective value of the service update, by the non-negativity of loss and $M$ we have that $h^k_j$ is a best-response to $M^{k+1}$.  Sticky updating gives that $h_j^{k+1} = h_j^k$. 

    Given a user $i$ at update $k > 0$, let us denote $A_i^k$ as the best response to services $H$.
    $A^{k+1}_i$ will be a user best response to $H$; since no other variables are involved in the user objective function, this is equivalent to choosing a new sample $A^k_i$ from the equivalence class of usages that maximizes the user objective function on $H$.  

    Due to the separability of the joint user and service updates, these asynchronous updates can be reanalyzed as parts of the joint update.
    As such, we can collapse all consecutive user updates and all consecutive service updates; without loss of generality, rounds can now be seen as a series of alternating user and service updates.  This can be re-indexed as a series of timesteps, each composed of one joint user and one joint service update. 

    By Theorem~\ref{thm:main}, this concludes the proof.
\end{proof*}

\section{ADDITIONAL EXPERIMENTS}

We present additional experiments on the settings introduced in Section~\ref{sec:experiments}.

\begin{figure*}[h]
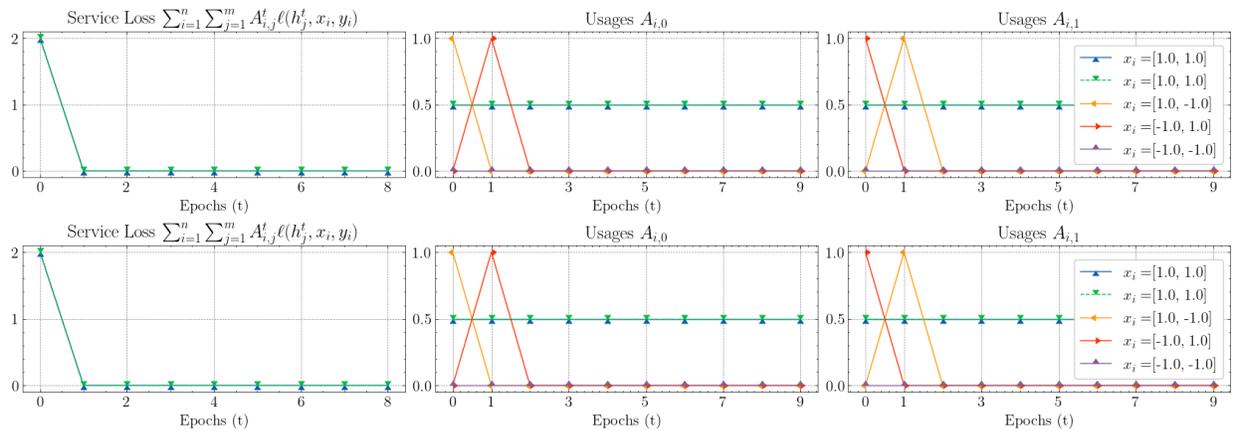

    \begin{center}
        \includegraphics[width=.95\textwidth]{supp_graphics/5_p1_m2.png} \\
        \includegraphics[width=.95\textwidth]{supp_graphics/5_p10_m2.png}
    \end{center}
    \caption{5-Points dataset; the top three graphs give the $p=0.1$ case while the bottom three give $p=1.0$.  }
    \label{fig:5_point_supp}
\end{figure*}
\paragraph{Synthetic Dataset} In Figure~\ref{fig:5_point_supp}, we show that different values of $p > 0$ don't affect the convergence.  The top set of plots gives service loss and usages for $p=0.1$ and the bottom set gives the same for $p=1$; however, both values of $p$ give the same usages and losses as each other across epochs. 

\begin{figure*}[h]
    \begin{center}
        \includegraphics[width=.24\textwidth]{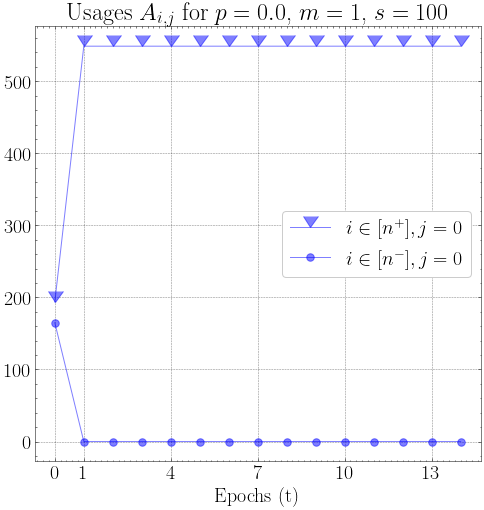} 
        \includegraphics[width=.24\textwidth]{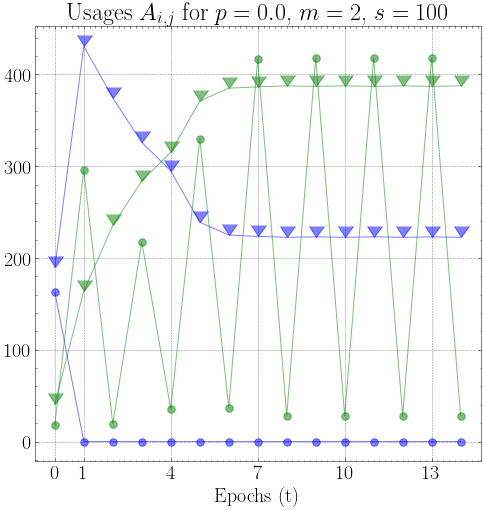}
        \includegraphics[width=.24\textwidth]{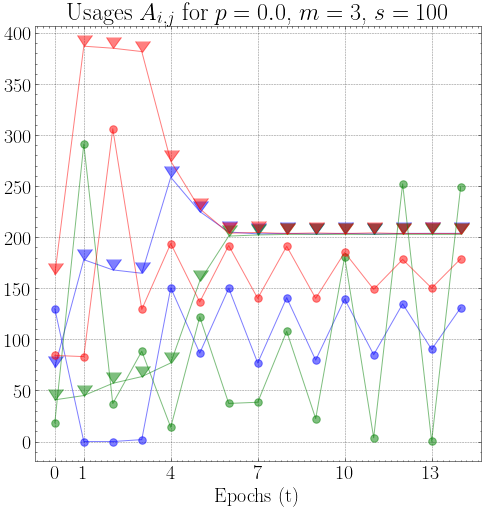} 
        \includegraphics[width=.24\textwidth]{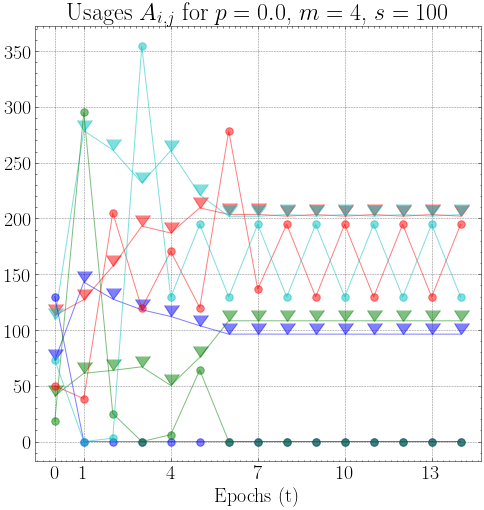} \\
        \includegraphics[width=.24\textwidth]{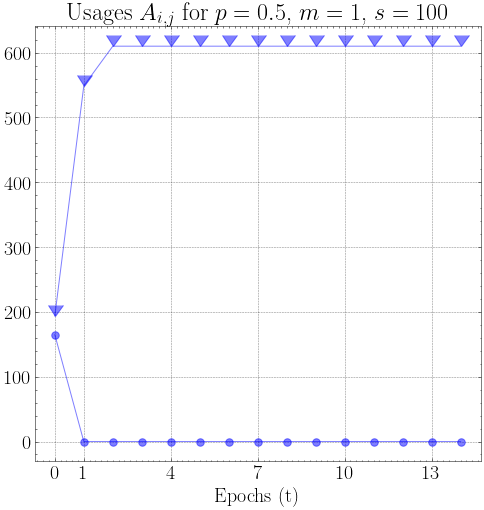} 
        \includegraphics[width=.24\textwidth]{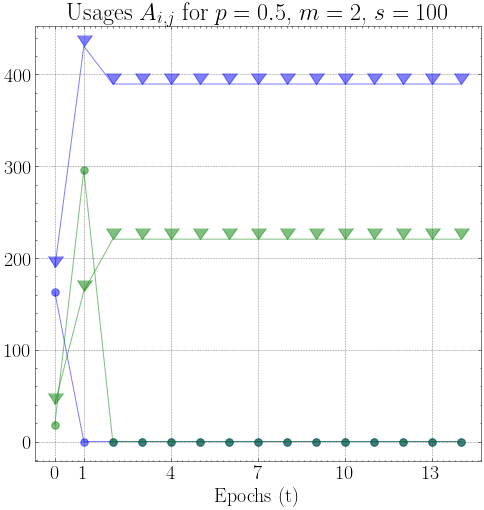}
        \includegraphics[width=.24\textwidth]{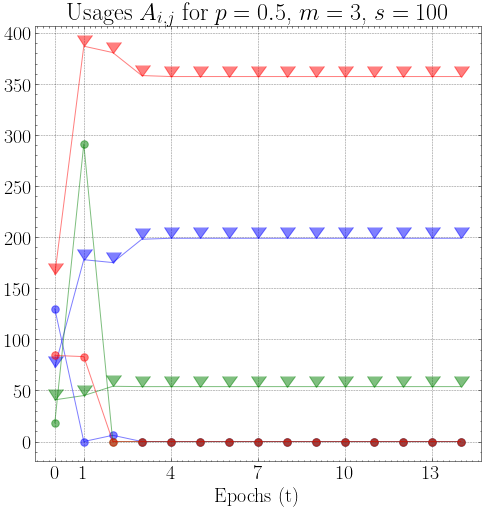} 
        \includegraphics[width=.24\textwidth]{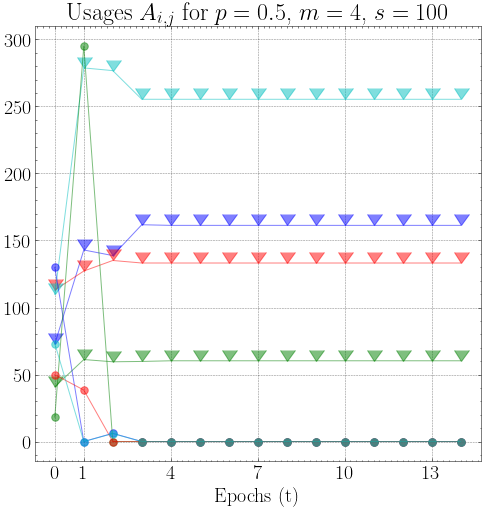}
    \end{center}
    \caption{Banknote Authentication dataset; ablating on $m$.  The top four graphs give the $p=0$ case while the bottom four give $p=0.5$.  
    In each graph, triangle markers indicate positive usages while circular markers indicate negative usages; colors indicate the different services.}
    \label{fig:banknote_m}
\end{figure*}
\paragraph{Banknote Authentication} Figure~\ref{fig:banknote_m} demonstrates how varying numbers of services can affect convergence.  Plots for $m=1, 2, 3, 4$ are given, both in the zero memory and nonzero memory cases.  This illustrates that as the number of services increases, convergence may take longer due to services interfering with each other and disincentivizing users to reveal themselves to other services through usage.  
Note that due to the static seed, between graphs models are shown the same initial users when the model is present.

\begin{figure*}[h]
    \begin{center}
        \includegraphics[width=.24\textwidth]{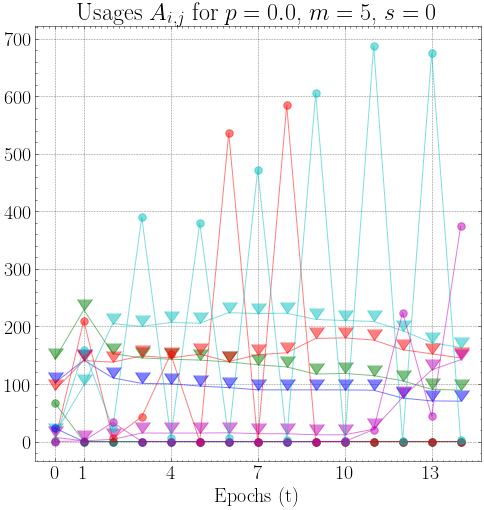} 
        \includegraphics[width=.24\textwidth]{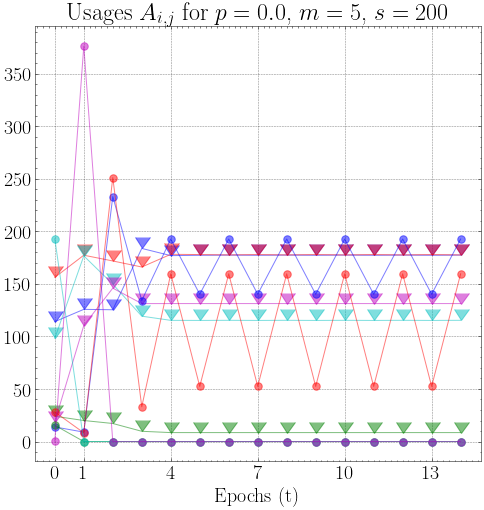}
        \includegraphics[width=.24\textwidth]{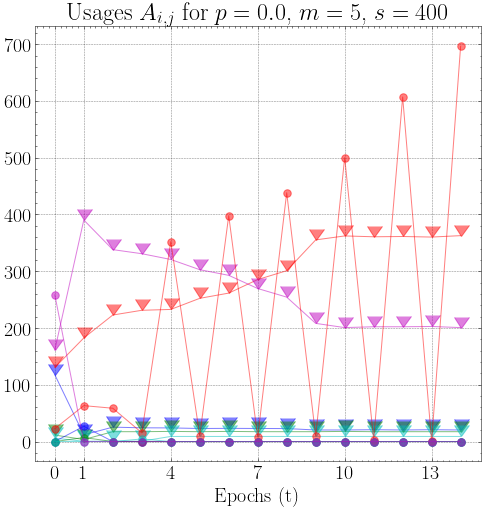} 
        \includegraphics[width=.24\textwidth]{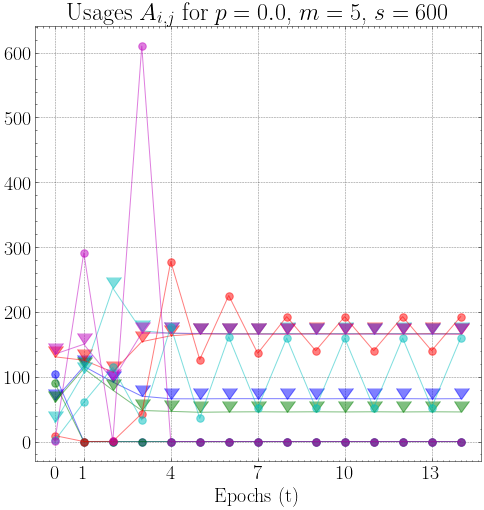} \\
        \includegraphics[width=.24\textwidth]{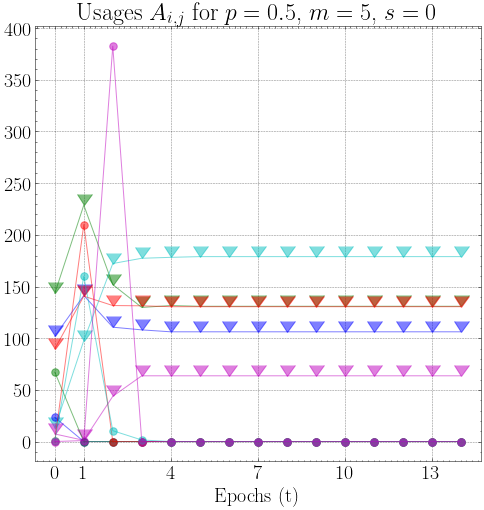} 
        \includegraphics[width=.24\textwidth]{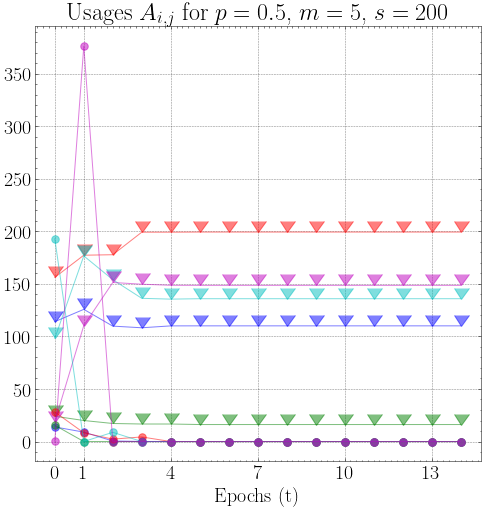}
        \includegraphics[width=.24\textwidth]{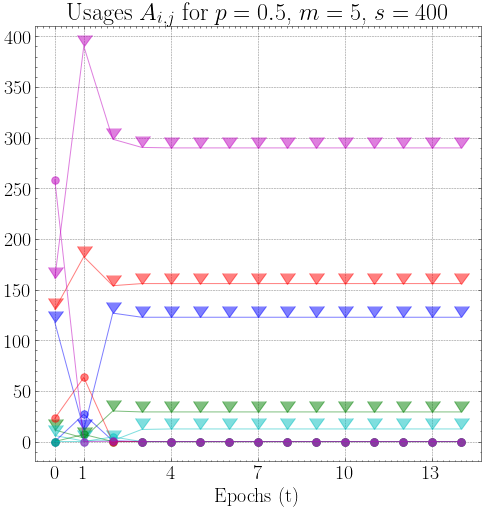} 
        \includegraphics[width=.24\textwidth]{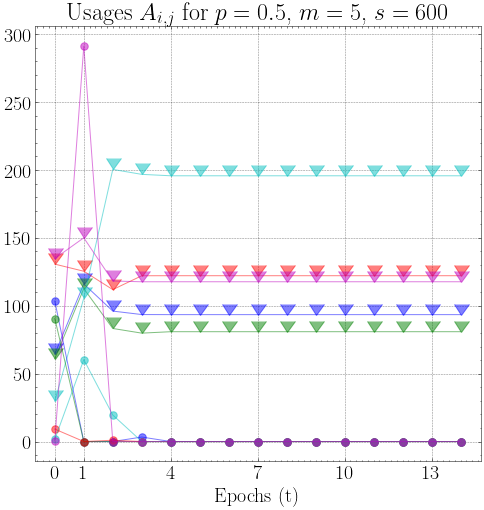}
    \end{center}
    \caption{Banknote Authentication dataset; ablating on the seed ($s$).  The top four graphs give the $p=0$ case while the bottom four give $p=0.5$. In each graph, triangle markers indicate positive usages while circular markers indicate negative usages; colors indicate the different services.}
    \label{fig:banknote_seed}
\end{figure*}
We illustrate the potential for a variety of outcomes depending on the initial seed in Figure~\ref{fig:banknote_seed}.  Generally, this shows that the initial conditions can have drastic effects on both the time to convergence and the final stable state.

\begin{figure*}[h]
    \begin{center}
        \includegraphics[width=.24\textwidth]{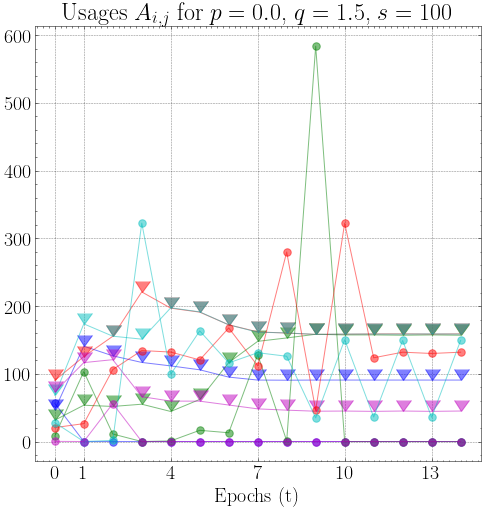} 
        \includegraphics[width=.24\textwidth]{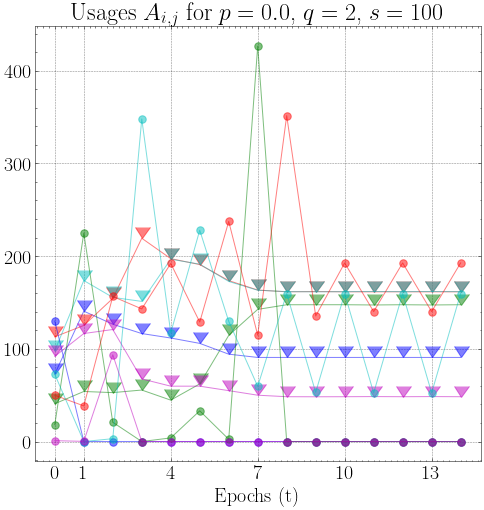} \includegraphics[width=.24\textwidth]{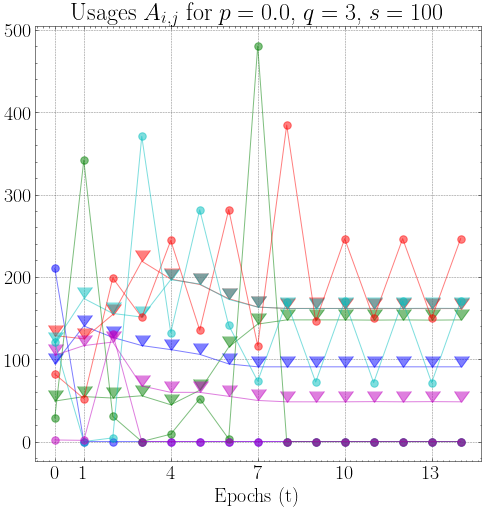} \includegraphics[width=.24\textwidth]{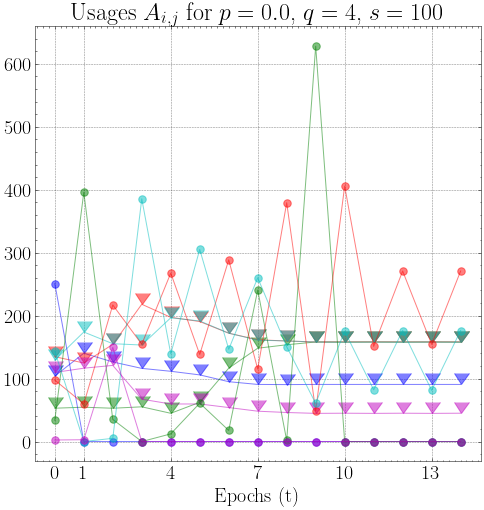} \\
        \includegraphics[width=.24\textwidth]{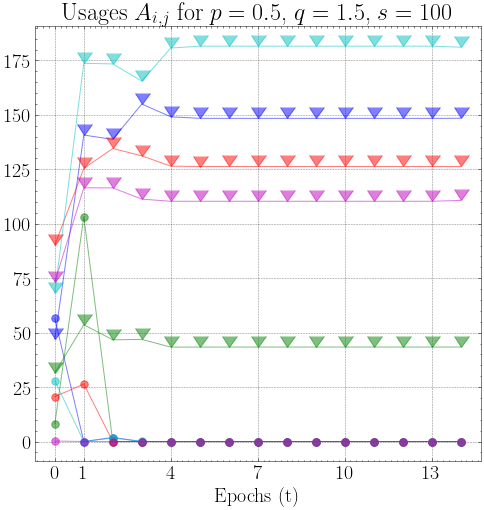} \includegraphics[width=.24\textwidth]{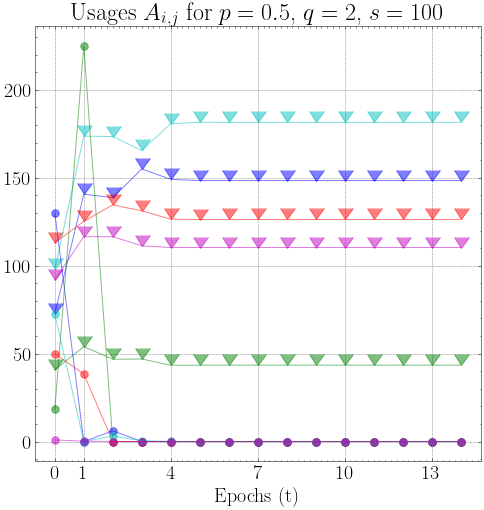} \includegraphics[width=.24\textwidth]{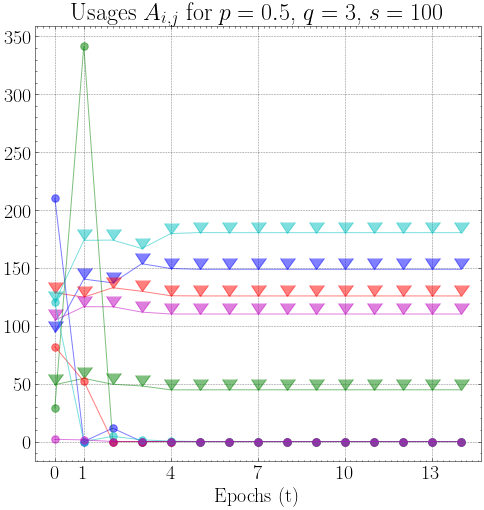} \includegraphics[width=.24\textwidth]{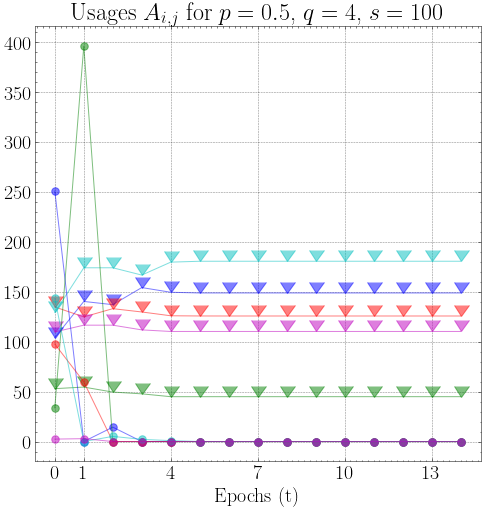} 
    \end{center}
    \caption{Banknote Authentication dataset; ablating on usage cost power factor $q$.  The top four graphs give the $p=0$ case while the bottom four give $p=0.5$. In each graph, triangle markers indicate positive usages while circular markers indicate negative usages; colors indicate the different services.}
    \label{fig:banknote_q}
\end{figure*}
To inspect the effect of cost power factor $q$ on convergence, we provide usages over epochs for varying values of $q$ both with and without memory in Figure~\ref{fig:banknote_q}.
No meaningful relationship with convergence is found for $q > 1$.

\paragraph{Bank Account Fraud} All experiments were run for $m=5$ services, with cost power factor $q=2$.  From the realizable subset of the data, we select the first $5,000$ positive and $5,000$ negative points to speed up runtime.

\begin{figure*}[h]
    \begin{center}
        \includegraphics[width=.24\textwidth]{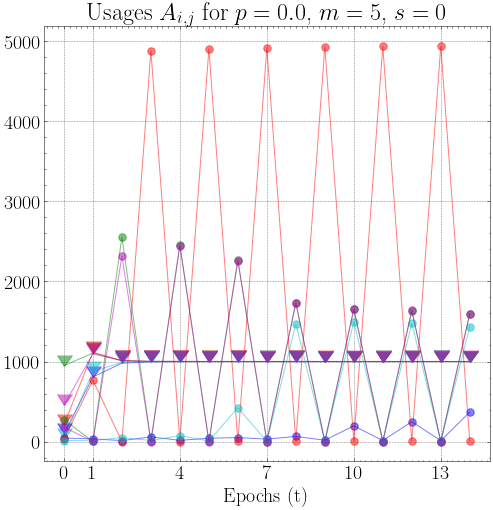} 
        \includegraphics[width=.24\textwidth]{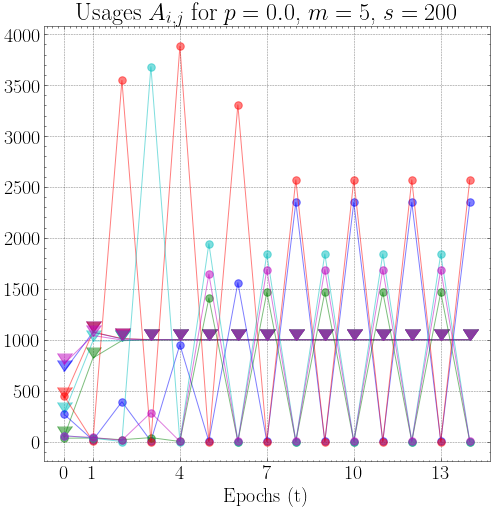}
        \includegraphics[width=.24\textwidth]{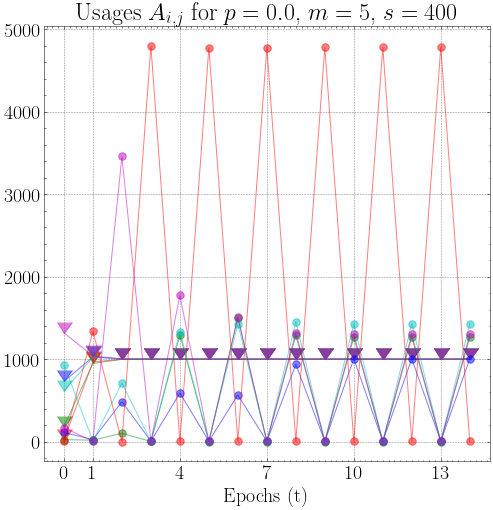} 
        \includegraphics[width=.24\textwidth]{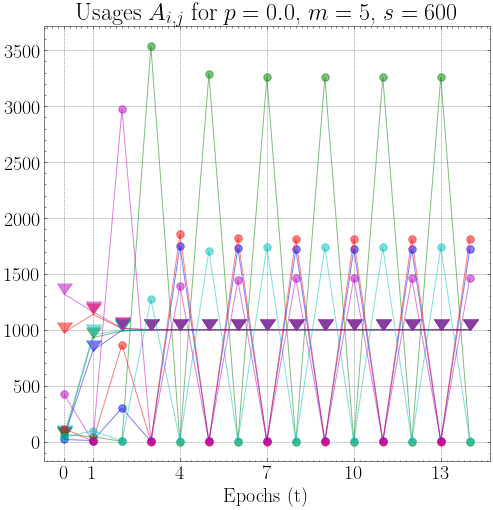} \\
        \includegraphics[width=.24\textwidth]{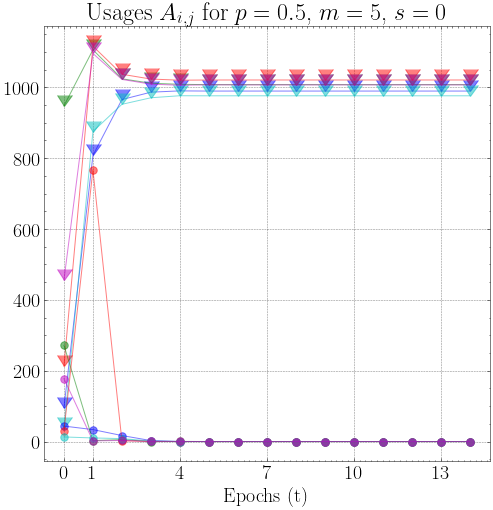} 
        \includegraphics[width=.24\textwidth]{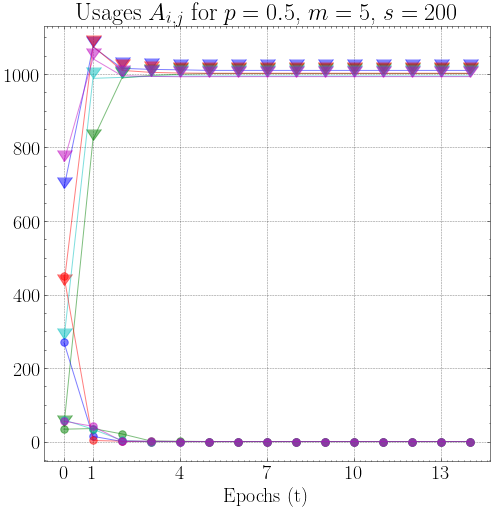}
        \includegraphics[width=.24\textwidth]{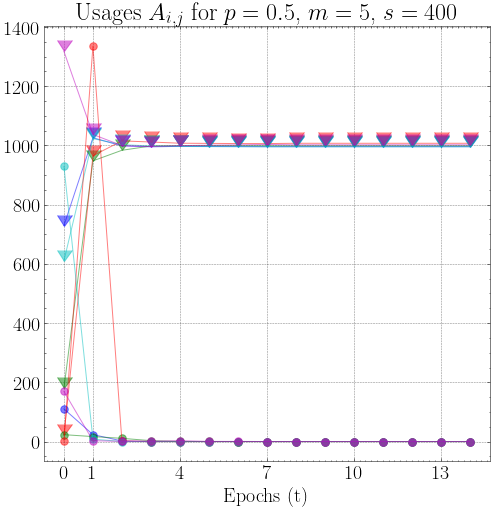} 
        \includegraphics[width=.24\textwidth]{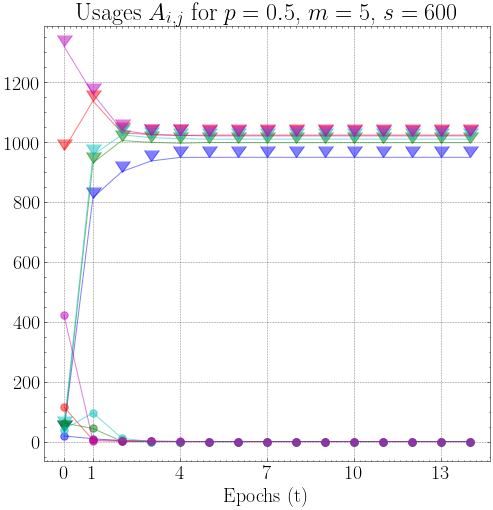}
    \end{center}
    \caption{Bank Account Fraud dataset; ablating on the seed ($s$).  The top four graphs give the $p=0$ case while the bottom four give $p=0.5$. In each graph, triangle markers indicate positive usages while circular markers indicate negative usages; colors indicate the different services.}
    \label{fig:baf_seed}
\end{figure*}
Results under a variety of initial seeds can be seen in Figure~\ref{fig:baf_seed}, including both results with and without memory.  These results further corroborate the earlier theoretical findings, demonstrating convergence in the memory case with oscillation commonly occurring in the na\"ive retraining setting.

\subsection{Hardware and Specifications}

All experiments were run on an Intel(R) Core(TM) i9-10885H CPU @ 2.40GHz.  
Our code is available at \url{https://github.com/eliotshekhtman/strategic-usage}.

\end{document}